\documentclass[%
endfloats*,
preprint,
nofootinbib,
amsmath,amssymb,
aps,
floatfix,
]{revtex4-1}

\usepackage{graphicx} 
\graphicspath{ {Figures/} }
\usepackage{epstopdf}
\usepackage[caption=false]{subfig}
\usepackage{amsmath}
\usepackage{amsfonts}
\usepackage{amssymb}
\usepackage{bm}
\usepackage{hyperref}
\usepackage{amsthm}
\usepackage{mathrsfs}
\usepackage[vlined,ruled]{algorithm2e}
\usepackage{url}
\usepackage{color}
\usepackage{algorithmic}
\usepackage{bbm}
\usepackage{booktabs}
\usepackage{array}
\usepackage[table]{xcolor}
\usepackage{yfonts}
\usepackage{lineno}
\usepackage{mathdots}

\usepackage{upgreek}

\usepackage{lineno}

\newcommand{\paren}[1]{\left( #1 \right)}
\newcommand{\brackets}[1]{\left[ #1 \right]}
\newcommand{\braces}[1]{\left\{ #1 \right\}}
\newcommand{\norm}[1]{\lVert #1 \rVert}

\newcommand{\bmrm}[1]{\bm{\mathrm{ #1 }}}

\newcommand{\map}[3]{#1: #2 \rightarrow #3}

\newcommand{\until}[1]{\braces{1,\dots, #1}}

\newcommand{\integer}{\mathbb{Z}}

\newcommand{\methods}{Methods}

\newcommand{\argmin}[2] {\mathrm{arg}\min_{#1}#2}

\DeclareSymbolFont{bbold}{U}{bbold}{m}{n}
\DeclareSymbolFontAlphabet{\mathbbold}{bbold}

\newcommand*{\rom}[1]{\expandafter\@slowromancap\romannumeral #1@}

\begin{document}


\bibliographystyle{naturemag}
\title{Characterising User Transfer Amid Industrial Resource Variation: A Bayesian Nonparametric Approach}

\author{Dongxu Lei}
\affiliation{Research Institute of Intelligent Control and Systems, Harbin Institute of Technology, Harbin, China}

\author{Xiaotian Lin}
\affiliation{Intelligent Control and Systems Research Center, Yongjiang Laboratory, Ningbo, China }

\author{Xinghu Yu}
\affiliation{Ningbo Institute of Intelligent Equipment Technology Company, Ltd., Ningbo, China}

\author{Zhan Li}
\affiliation{Research Institute of Intelligent Control and Systems, Harbin Institute of Technology, Harbin, China }

\author{Weichao Sun}
\affiliation{Research Institute of Intelligent Control and Systems, Harbin Institute of Technology, Harbin, China  }

\author{Jianbin Qiu}
\affiliation{Research Institute of Intelligent Control and Systems, Harbin Institute of Technology, Harbin, China   }

\author{Songlin Zhuang}
\affiliation{Intelligent Control and Systems Research Center, Yongjiang Laboratory, Ningbo, China}

\author{Huijun Gao}
\affiliation{Research Institute of Intelligent Control and Systems, Harbin Institution of Technology, Harbin, China    \\
Intelligent Control and Systems Research Center, Yongjiang Laboratory, Ningbo, China\\ To whom correspondence should be addressed: songlin-zhuang@ylab.ac.cn, hjgao@hit.edu.cn
}

\date{\today}

\clearpage

\ \\

\begin{abstract}

In a multitude of industrial fields, a key objective entails optimising resource management whilst satisfying user requirements. Resource management by industrial practitioners can result in a passive transfer of user loads across resource providers, a phenomenon whose accurate characterisation is both challenging and crucial. This research reveals the existence of user clusters, which capture macro-level user transfer patterns amid resource variation. We then propose CLUSTER, an interpretable hierarchical Bayesian nonparametric model capable of automating cluster identification, and thereby predicting user transfer in response to resource variation. Furthermore, CLUSTER facilitates uncertainty quantification for further reliable decision-making. Our method enables privacy protection by functioning independently of personally identifiable information. Experiments with simulated and real-world data from the communications industry reveal a pronounced alignment between prediction results and empirical observations across a spectrum of resource management scenarios. This research establishes a solid groundwork for advancing resource management strategy development.

\end{abstract}

\maketitle


\section{Introduction}

In diverse industrial fields, a fundamental goal is to optimise resource management to boost profitability and sustainability whilst satisfying user requirements. Central to this process is the manipulation of resource availability (RA) by industrial practitioners amongst resource providers (RPs). RA indicates the accessibility of vital resources—such as energy, materials, or infrastructure—at the moment of necessity, whilst RPs refer to entities that supply these resources. Beyond direct manipulation, RA can also be influenced by external factors, including market dynamics, geographical and temporal fluctuations, policy regulations, etc. User load (UL) refers to the volume or intensity of demand exerted by users on RPs. Within this context, users respond to RA variations by seeking and transitioning towards more viable alternatives, resulting in passive transfer amongst RPs and varying ULs, as in Fig. \ref{fig:1}(a)-(b). Effective resource management entails predicting user transfer amid resource variations. Therefore, characterising this transfer phenomenon contributes significantly to achieving the goal of resource optimisation and ensuring economic viability.

The issue of resource optimisation has emerged as a focal point for both policymakers and scholars in recent years. For instance, in the energy field, the US and EU policymakers \cite{osti_1842610} have been prompted by political \cite{ruhnau2023natural} and economic factors to adopt demand response programmes. These programmes bring about commercial advantages \cite{lee2022datasets} and advocate the use of renewable resources \cite{alahaivala2017framework, kocaman2020stochastic}. Similarly, demand-responsive transport, a concept born in the 1970s, gained traction in Europe and North America \cite{mageean2003evaluation, ellis2009guidebook} by adjusting public transport routes to meet specific demands. Also, the communications industry has engaged in research into optimal resource allocation strategies \cite{fooladivanda2012joint, zhao2019deep, khalili2020joint, qin2018user}, all highlighting the importance of resource optimisation in maintaining industrial competitiveness. Whilst the development of effective and efficient resource management strategies critically relies on an accurate and reliable prediction model for user transfer in response to variations in RA, the establishment of such a model remains an open problem.

The model should fulfil two primary requirements: 1) accuracy and 2) prediction uncertainty quantification. This modelling task is often challenged by the limited quantity of available historical data across many realistic scenarios. For instance, in communication systems, the operational status of certain base stations (BSs) seldom changes. This limited variability in the dataset can engender extrapolation errors, potentially undermining the model's capacity for accurate prediction. In many industrial fields, data anonymisation requirements, mandated for the preservation of privacy \cite{lee2022datasets, girka2021anonymization}, introduce an additional layer of complexity to modelling. This process necessitates the removal of personally identifiable information from historical datasets, thereby limiting the informational resources accessible to RP operators. Such a paucity of comprehensive information contributes to uncertainties in predicting ULs linked to RPs, a phenomenon corroborated extensively by existing research \cite{hunt2000modelling, yamaguchi2017stochastic, corman2021stochastic, ni2018modeling, jiang2016timegeo}. This prediction uncertainty disqualifies approaches that merely yield a point estimate, underscoring the preference for models that can effectively quantify the uncertainty \cite{van2019amplification}. 

The task of predicting ULs with the RA of RPs serving as explanatory variables can be construed as a regression problem in a supervised setting within the machine learning (ML) paradigm. Despite ML being extensively employed to model complex industrial processes in a data-driven manner \cite{bishop2006pattern, jordan2015machine, brunton2022data, albora2023product}, several drawbacks render out-of-the-box ML techniques impractical for addressing our specific problem. Highly predictive ML models such as convolutional neural networks \cite{gu2018recent, albawi2017understanding, aghdam2017guide, tan2019efficientnet, kiranyaz20211d} and recurrent neural networks \cite{medsker1999recurrent, yu2019review, graves2007multi, hermans2013training, mandic2001recurrent} are hampered by the lack of interpretability, which presents a significant obstacle in the context of industrial decision-making \cite{chen2023physics, nakamura2021health}. Techniques like traditional statistical analysis and linear models are more interpretable, yet the accuracy of these approaches is often found wanting. Whilst methodologies like Gaussian process regression \cite{schulz2018tutorial, deringer2021gaussian, banerjee2013efficient, williams1995gaussian, bernardo1998regression} and specific neural network methodologies \cite{kasiviswanathan2017methods, gawlikowski2021survey, eaton2018towards, qiu2019quantifying, quan2014uncertainty, pierce2008uncertainty, kasiviswanathan2016quantification, chitsazan2015prediction} can handle uncertainty, their application is hindered by their black-box nature and the challenges posed by extrapolation error. To the best of our knowledge, no modelling method has been developed that can accurately, interpretably, and reliably characterise the user transfer process.

In this study, we introduce CLUSTER (Characterising Latent User Structure Through Evidence Refinement), a hierarchical Bayesian nonparametric model customised to analyse interactions between user preferences (UPs) and the RA of each RP, which together influence the observed ULs, as in Fig. \ref{fig:1}(c). CLUSTER capitalises on the inherent clustering of users with similar UPs to predict ULs. Two versions of CLUSTER are implemented: Na\"ive CLUSTER, which requires manual cluster number specification, and Complete CLUSTER, which uses a Dirichlet process mixture model (DPMM) to autonomously determine cluster numbers, enhancing model flexibility and reducing computational burden. CLUSTER generates a posterior predictive distribution for the ULs associated with each RP, furnishing a comprehensive representation of prediction uncertainty. Additionally, specifically designed to work with anonymised data, CLUSTER requires only aggregate data, such as the average or sum of ULs from the RP side, thus respecting privacy by eliminating the need for personally identifiable information. We evaluate CLUSTER using both simulated and real-world communications datasets. Na\"ive CLUSTER is tested for inferring UPs and clustering, whilst Complete CLUSTER demonstrates flexibility in automatically grouping users with similar UPs. A comprehensive assessment confirms CLUSTER's accuracy and ability to quantify uncertainty. In both simulated and real-world settings, experimental findings demonstrate a significant alignment between predictive outcomes and empirical data across diverse resource management contexts, validating CLUSTER's effectiveness in managing large-scale and complex scenarios. The work contributes to the existing literature on model-based resource optimisation.

\newpage
\section{Results}

\subsection{Probabilistic Analysis of User Transfer Dynamics}

We begin with a mathematical description of the problem. Consider a system governed by the following unknown deterministic dynamics \(f\),
\begin{linenomath*}
\begin{equation}\label{eq:sys}
    \bmrm{x}\paren{k + 1} = f\paren{\bmrm{x}\paren{k}, \bmrm{u}\paren{k}, \bm{\upeta}\paren{k}},
\end{equation}
\end{linenomath*}
where \(N \in \integer\), \(\bmrm{x}\paren{k} \in \left[0, +\infty\right)^{N}\) and \(\bmrm{u}\paren{k} \in \left[0, +\infty\right)^{N}\), denote respectively, the number of RPs, the ULs (response variable) associated each RP and the corresponding RA (explanatory variable). The term \(\bm{\upeta}\paren{k}\) indicates unidentified influencing factors, whose exact interpretation and dimensionality are usually elusive. Across varying applications, \(\bmrm{u}\paren{k}\) is often subject to human adjustment, allowing for manipulation of the represented resources. The variables \(\bmrm{x}\paren{k}\) and \(\bmrm{u}\paren{k}\) can manifest as either aggregate or instantaneous data, as in Fig. \ref{fig:2}(a). This study focuses mainly on aggregate data, as it aligns with the real-world data utilised in Section \ref{sec:Validation on realistic data}. Nevertheless, our proposed framework retains the flexibility to accommodate instantaneous applications and extends seamlessly to such cases. 

The objective of our research is to predict ULs, \(\bmrm{x}\paren{k + 1}\), subsequent to variations in RA, \(\bmrm{u}\paren{k}\), at time \(k\). The term \(\bm{\upeta}\paren{k}\) introduces uncertainty to the prediction of \(\bmrm{x}\paren{k + 1}\), rendering deterministic prediction of \(\bmrm{x}\paren{k + 1}\) infeasible. Consequently, we formalise \(\bmrm{x}\paren{k + 1}\) as a random vector whose conditional probability distribution is partially determined by \(\bmrm{u}\paren{k}\). In our efforts to quantify the uncertainty associated with \(\bmrm{x}\paren{k + 1}\), we utilise a customised probabilistic model, termed CLUSTER and denoted by \(\mathcal{M}\). This model leverages the following posterior predictive distribution:
\begin{linenomath*}
\begin{equation}\label{eq:post_pred}
    p\paren{\bmrm{x}\paren{k + 1} | \bmrm{x}\paren{k}, \bmrm{u}\paren{k}, \mathcal{D}, \mathcal{M}} 
    = \int_{\bm{\Theta}} p\paren{\bmrm{x}\paren{k + 1} | \bmrm{x}\paren{k}, \bmrm{u}\paren{k}, \bm{\uptheta}, \mathcal{M}} p\paren{\bm{\uptheta} | \mathcal{D}, \mathcal{M}} \mathrm{d} \bm{\uptheta},
\end{equation}
\end{linenomath*}
which enables the prediction for \(\bmrm{x}\paren{k + 1}\) under an altered \(\bmrm{u}\paren{k}\), considering the historically observed dataset \(\mathcal{D}\) and the latent variables \(\bm{\uptheta}\). Here, \( \bm{\Theta} \) delineates the latent variable space and \(\mathcal{D} = \left\{\bmrm{x}\paren{k_{i} + 1}, \bmrm{u}\paren{k_{i}}\right\}_{i=1}^{D}\), where \(D\) signifies the size of the historical data, as in Fig. \ref{fig:2}(a). The components encompassed within \(\bm{\uptheta}\) and the structure of \(\mathcal{M}\) are elucidated in Section \ref{sec:modeling}.

\subsection{Mitigating Extrapolation Errors via User Preference Parameterisation}\label{sec:user_RA}

Modelling the posterior predictive distribution in Equation \eqref{eq:post_pred} is challenging due to the high dimensionality of \(\bmrm{x}\paren{k + 1}\) and \(\bmrm{u}\paren{k}\), and the need for large training datasets, especially in industrial settings with numerous RPs exhibiting limited RA variability. To mitigate these challenges, we introduce a new latent variable, UP, for each individual user to simplify the modelling process. To elucidate, understood as the probability of a user being associated with each RP when all RPs are available, UP serves as an inherent characteristic for each user, signifying their inclination for specific RPs. In addition to accommodating scenarios wherein users opt for RPs in a stochastic manner, our methodology is also applicable to instances involving deterministic selection. Further elucidation is provided in Supplementary Note \uppercase\expandafter{\romannumeral5}. This attribute is shaped by a myriad of factors, such as socio-economic status, geographical location, and behavioural psychology, and is independent of the current RA conditions. Our approach is predicated on the observation that the ULs are determined by a complex interplay between the UPs and the RA, as in Fig. \ref{fig:1}(c). 

We first model the latent variables of CLUSTER by considering the problem involving \(M\) users and define \(\bmrm{l}^{j}\) as the UP for the \(j\)-th user, where \(M \in \mathbb{N}\), \(j \in \until{M}\), and \(\bmrm{l}^{j} \in \Delta^{N-1}\), with \(\Delta^{N-1}\) representing the standard \(\paren{N-1}\)-simplex. As a result, the value of each component of \(\bmrm{l}^{j}\) can be construed as the probability of a user associated with the corresponding RP, given the full availability of all RPs. The UPs amongst various users exhibit pronounced heterogeneity. In the absence of prior knowledge regarding this diversity, we regard the UPs of distinct users as exchangeable multivariate random variables. 

In our efforts to analyse the influence of \(\bmrm{u}\paren{k}\) on \(\bmrm{x}\paren{k + 1}\), we establish a metric known as the preference score (PS) for each user, given by \(\bmrm{s}^{j}\paren{k + 1} := h\paren{\bmrm{l}^{j}, \bmrm{u}\paren{k}}\), where \(\bmrm{s}^{j}\paren{k + 1} \in \Delta^{N-1}\). The function \(\map{h}{\Delta^{N-1} \times \left[0, +\infty\right)^{N}}{\Delta^{N-1}}\) acts as the preference score function (PSF), quantifying the combined impact of users' UPs and RPs' RA on users' eventual associations. The advancement of one time step signifies the temporal delay in the influence of \(\bmrm{u}\paren{k}\) upon \(\bmrm{x}\paren{k + 1}\). The PSs are constrained to reside in \(\Delta^{N-1}\). Consequently, this PS offers a perspective on the probability associated with a user's selection of each RP in light of the current RA. As a design choice, we opt for a straightforward and insightful PSF:
\begin{linenomath*}
\begin{equation}\label{eq:h_func}
h\paren{\bmrm{l}^{j}, \bmrm{u}\paren{k}} := \frac{\bmrm{l}^{j} \odot \bmrm{u}\paren{k}}{\bmrm{l}^{j} \cdot \bmrm{u}\paren{k}},
\end{equation}
\end{linenomath*}
where the operation \(\odot\) signifies the element-wise multiplication between two vectors. The selection of \(h\) simply reflects the collective contribution of the UPs and the RA of RPs.

Users exhibiting similar UPs manifest a natural tendency to aggregate into clusters, evidence for which is furnished in the subsequent section. This phenomenon enables analysis at the macro level of the UP for each cluster, thus considerably reducing the scale of model parameters and driving computational efficiency. In a scenario where the \(M\) users aggregate into \(W\) clusters, with each cluster sharing a common PS and distinct clusters having different PSs, we expand our analysis to consider this clustering framework. Associated with the clusters is a weight vector \(\bmrm{w} = \brackets{\mathrm{w}_{1}, \ldots, \mathrm{w}_{W}}^{\top}\), where \(\bmrm{w} \in \Delta^{W - 1}\). We introduce the proportion vector \(\bmrm{p}\paren{k + 1}\), which stands for the mathematical expectation of the proportion of ULs associated with each RP, derived from a linear combination of the PSs and the weights corresponding to each cluster. Rigorous mathematical derivation leads to the following expression for \(\bmrm{p}\paren{k + 1}\):
\begin{linenomath*}
\begin{equation}\label{eq:p}
\bmrm{p}\paren{k + 1} = \paren{\operatorname{diag}\paren{\bmrm{L} \operatorname{diag}\paren{\bmrm{u}\paren{k}} \bmrm{1}_{N}}^{-1} \bmrm{L} \operatorname{diag}\paren{\bmrm{u}\paren{k}}}^{\top} \bmrm{w},
\end{equation}
\end{linenomath*}
where \(\bmrm{L} = \brackets{\bmrm{l}^{1}, \ldots, \bmrm{l}^{W}}^{\top}\), and \(\operatorname{diag}\paren{\bmrm{v}}\) denotes a diagonal matrix with entries from a vector \(\bmrm{v}\). The detailed derivation can be found in the Supplementary Note \uppercase\expandafter{\romannumeral3}.

For the final stage of likelihood modelling, our method deploys the Dirichlet distribution, supplemented by a positive latent concentration variable \(\mathrm{c}\), to jointly model the mean and variance of the UL proportions associated with each RP. Mathematically, this relationship is defined as:
\begin{linenomath*}
\begin{equation}\label{eq:proportion_likelihood}
\frac{\bmrm{x}\paren{k + 1}}{M} \sim \operatorname{Dirichlet}\paren{\mathrm{c}\bmrm{p}\paren{k + 1}},
\end{equation}
\end{linenomath*}
where \(M\) denotes the overall quantity of ULs across all RPs. Overall ULs generally remain invariant in the face of RA variations and can be captured by \(M = \norm{\bmrm{x}\paren{k}}_{1}\), where \(\norm{\cdot}_{1}\) signifies the \(\ell^{1}\) norm. The necessity for independent variance modelling and the superiority of our choice over other alternatives like the multinomial distribution are detailed in Supplementary Note \uppercase\expandafter{\romannumeral1}.

By employing UP parameterisation, we incorporate our prior knowledge about the problem into CLUSTER. This approach mitigates the extrapolation errors stemming from sparse data, capitalising on the distinct structural aspects of the problem. Furthermore, our method relies solely on anonymous aggregate data from the RP side, safeguarding confidential user details.

\subsection{User Preference Analysis Using CLUSTER}\label{sec:modeling}

We move forward to the inference of the latent variables within CLUSTER in this section. First, Na\"ive CLUSTER is employed to demonstrate the capability of CLUSTER to infer the UPs. Subsequently, we establish that the underlying structure of UPs provides valuable insights into the existence of user clusters, motivating the integration of the DPMM to facilitate automated clustering. The integration of the DPMM in Complete CLUSTER enhances flexibility and further reduces the scale of model parameters, minimising the potential error that might be introduced by the heuristic decisions of the model designer on cluster numbers.

\subsubsection{Revelation of User Clusters: Simulated Dataset Analysis via Na\"ive CLUSTER}\label{sec:naive_model}

We start by employing a simulated dataset obtained based on an example of wireless communications. In this context, mobile devices possess the capability to receive signals transmitted by multiple BSs located in proximity. The device selects the BS to establish a connection based on the received signal strength. On the supply side, BSs report the average number of users at fixed intervals, and for user privacy, only aggregate data are available, with identities and locations of individual users connected to each BS undisclosed. In this analysis, we create a simulated dataset with 10 BSs and 100 users. During the simulation, each BS alternates between `on' and `off' states. Afterwards, the entire simulation duration is divided into uniform intervals, recording both the proportion of active BS time and the average user count per BS (details in \methods).

\begin{figure*}[t]  


\centering   \includegraphics[width=0.925\linewidth]{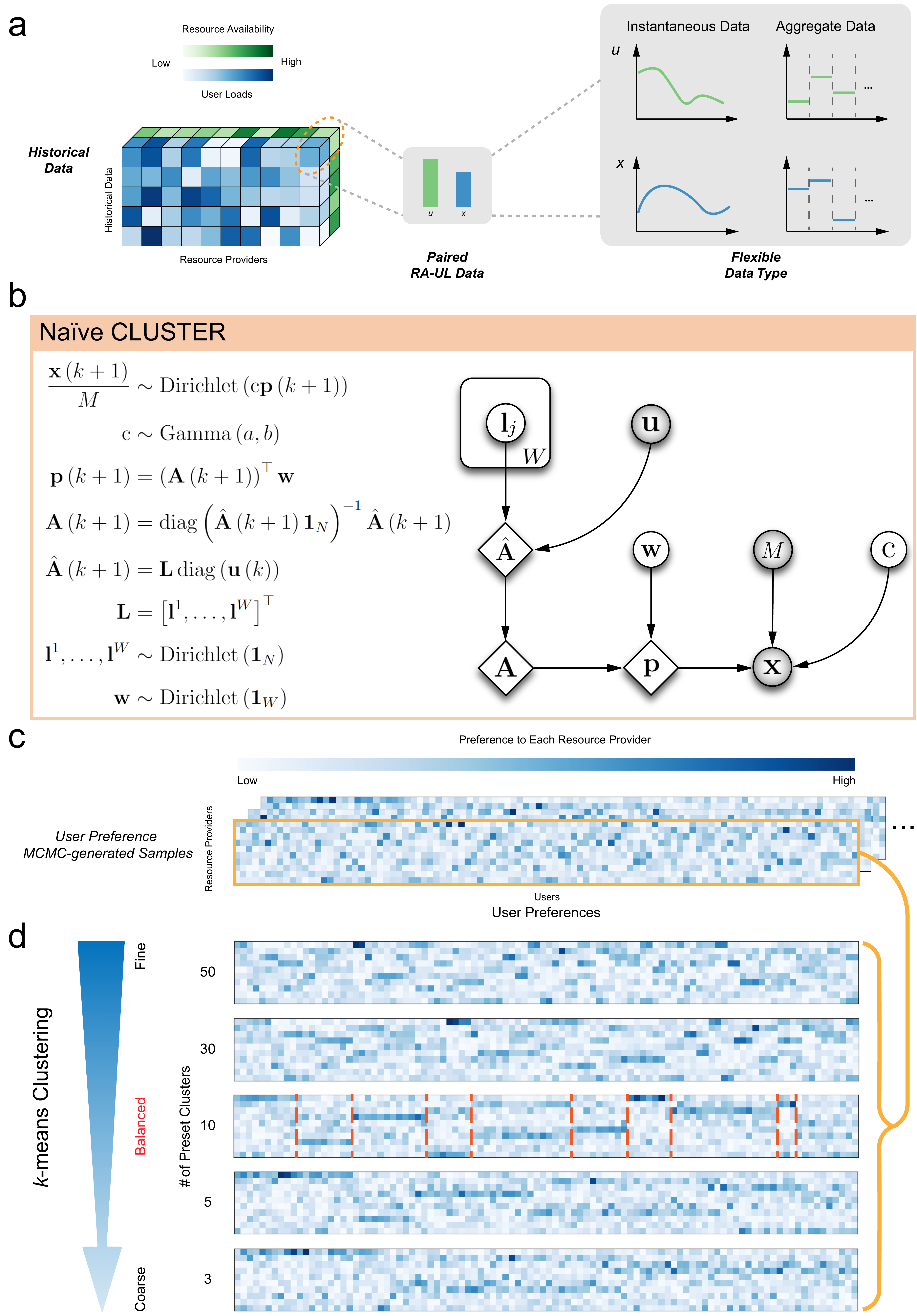}  \vspace{0.5cm} \caption{\linespread{1}\selectfont{}\textbf{Revealing User Clusters with Na\"ive CLUSTER.} In panel (\textbf{a}), the figure shows the architecture of the historical dataset used for this study, which accommodates both instantaneous and aggregate data types. The data pairs of RA and ULs represented as \(\bmrm{u}\paren{k}\) and \(\bmrm{x}\paren{k + 1}\), are input to CLUSTER and are colour-coded in green and blue, respectively. Panel (\textbf{b}) provides a graphical model of Na\"ive CLUSTER. In this diagram, rectangles stand for replications, circles symbolise random variables, diamonds denote deterministic variables, and shaded circles indicate observable variables. The \(W\) specifies the pre-set number of clusters. In panel (\textbf{c}), MCMC samples of the posterior distribution for UPs are displayed, with each column corresponding to an individual user. In this specific experiment, 100 clusters are pre-set to match the number of users. Variations in pixel shading signify relative preferences to corresponding RPs, with darker shades indicating higher preferences. Panel (\textbf{d}) uses \emph{k}-means clustering to analyse one posterior sample from panel (\textbf{c}). The study determines an optimal cluster number of 10, serving as a balanced midpoint between excessively fine and overly coarse representations. Orange vertical lines delineate partitions amongst user clusters.
}
\label{fig:2}
\end{figure*}

Fig. \ref{fig:2}(b) presents a graphical model delineating the relationships among the variables in our analysis and the detailed structure of Na\"ive CLUSTER. At this stage, we disable the DPMM component for illustrative purposes. Na\"ive CLUSTER posits \(W\) distinct user clusters, where the value of \(W\) is predetermined. Consequently, the latent variables within the Na\"ive CLUSTER are represented as \(\bm{\uptheta} = \braces{\bmrm{w}, \bmrm{L}, \mathrm{c}}\). Given our lack of prior information regarding UPs, we select noninformative priors, specifically \(\operatorname{Dirichlet}(\bmrm{1}_{N})\) for individual UPs \(\bmrm{l}^{j}\) and \(\operatorname{Dirichlet}(\bmrm{1}_{W})\) for \(\bmrm{w}\).

We utilise Markov Chain Monte Carlo (MCMC) methods to infer the posterior distribution of latent variables. This allows us to discern the structure of UPs from the MCMC-generated samples. For illustrative clarity, the number of clusters \(W\) is first set equal to the total number of users \(M\). As in Fig. \ref{fig:2}(c), each sample contains the UPs of the entire user population. Due to the exchangeability of UPs, we observe that the order of UPs for each user varies amongst the MCMC-generated posterior samples. However, the structure of the UPs can be discerned by manual clustering. The \(k\)-means clustering is applied to a representative sample. A notable clustering pattern emerges upon cluster reduction via \(k\)-means clustering, as shown in Fig. \ref{fig:2}(d). The results validate that users with similar UPs can aggregate into clusters in latent space. It is desirable to strike a balance between overly fine and overly coarse clusters.

In another experiment, we systematically adjust the preset number of clusters, \(W\), initiating the clustering process at the inference stage. Given that UPs naturally reside in a \(\paren{N - 1}\)-dimensional simplex, direct visualisation poses challenges. To address this, we employ t-distributed stochastic neighbour embedding (t-SNE) \cite{van2008visualizing} to project selected posterior samples of UPs into both 2-dimensional (Fig. \ref{fig:3}(a)) and 3-dimensional (see supplementary videos) representations. Detailed insights into this t-SNE mapping are provided in the Supplementary Note \uppercase\expandafter{\romannumeral6}. Through this visualisation, the existence of user clusters becomes apparent, underscoring the value of automating the determination of the number of clusters.

\subsubsection{Enhancing Model Flexibility: Integrating the DPMM for Automated Cluster Determination}\label{sec:complete_model}

The previous section emphasises the discernible clustering in the t-SNE mapping of posterior samples of UPs. Notably, Na\"ive CLUSTER operates under an assumption of \(M\) clusters. The manual determination of cluster number is dependent on the model designer's expertise, risking potential inaccuracies. Such a methodology constrains the model's flexibility and might misrepresent the inherent structure of the data. Conversely, an exhaustive search to ascertain the optimal cluster number would augment the computational demands. 

To address the constraints above, we incorporate the DPMM, obviating the need for \emph{a priori} specification of cluster number and mitigating the computational burden. The DPMM provides a more adaptive clustering mechanism, deducing the optimal number of clusters from the data. For this research, we utilise a stick-breaking process to facilitate DPMM implementation. The graphical representation and detailed structure of Complete CLUSTER can be found in Fig. \ref{fig:3}(b). Thus, the latent variables encompassed in Complete CLUSTER is \(\bm{\uptheta} = \braces{\upalpha, \bmrm{w}, \bmrm{L}, \mathrm{c}}\). Owing to the computational intractability of implementing DPMM's infinite cluster framework, CLUSTER employs a truncation scheme. Further specifications are available in \methods.

The integration of DPMM notably enhances the model's flexibility, effectively diminishing biases associated with arbitrary design choices. As evidenced in Fig. \ref{fig:3}(c), our analysis elucidates that a limited cluster count is sufficient to represent the dominant mixture weights, yielding a more compact data depiction. Consequently, this reduction in the number of clusters attenuates the computational requisites for the MCMC sampling processes.

\subsection{CLUSTER in Action: Posterior Predictive Analysis and Dataset Examination}

The posterior distribution of latent variables forms a crucial intermediate step within the CLUSTER workflow, bridging the gap between the historical user transfer and its future prediction, as illustrated in Fig. \ref{fig:4}(a). According to Equation \eqref{eq:post_pred}, it grants access to the posterior predictive distribution, which elucidates the user transfer under RA variation and serves as a crucial factor for accurate predictions and the quantification of uncertainties.

To evaluate CLUSTER's efficacy, posterior predictive checks are conducted on a testing dataset. The mathematical expectation of the predicted ULs for each RP constitutes the nominal prediction, which is compared against the ground truth to gauge nominal prediction error. Additionally, the standard deviation of these predictions offers insights into uncertainty quantification. Computational specifics are elaborated in \methods.

Whilst nominal prediction errors and standard deviations are straightforward metrics for assessing performance, their significance must be viewed in context with the inherent uncertainty of the observed random variables. A high error or deviation does not automatically indicate poor model performance, particularly for RPs with large variances in ULs. Thus, we employ the reliability plot, representing a calibration of our model \cite{gelman2013bayesian} (more details in \methods). This furnishes insights into the consistency between the model's predicted uncertainties and the actual empirical data. 

In the subsequent part of this section, we demonstrate the utility of CLUSTER through experiments conducted on both simulated and real-world datasets from the communications domain. These experiments underscore CLUSTER's capability to provide accurate predictions and effective uncertainty quantification. Refer to the Supplementary Figures for more details about the sampling results.

\subsubsection{Simulated Data Approach to CLUSTER's Prediction Analysis}\label{sec:valid_simulated}

In the evaluation of the predictive efficacy of the Complete CLUSTER model, posterior predictive checks are employed on a simulated dataset as delineated in Section \ref{sec:naive_model}. The multivariate nature of the response variables introduces analytical intricacies, further elaborated upon in the \methods~section. Considering the stochastic dynamics governing the user transfer process, absolute predictive accuracy in ULs remains an impractical goal. However, Fig. \ref{fig:4}(b) reveals an obvious aggregation of nominal prediction errors around zero, underscoring the model's general accuracy.

It is prudent to note that nominal prediction error alone does not offer a comprehensive appraisal of the model's performance. Higher inherent variability in the user transfer process can naturally lead to inflated nominal errors. To assess the variability, an analysis of prediction uncertainty is undertaken. Fig. \ref{fig:4}(c) showcases the standard deviations of prediction outcomes across various RPs, predominantly ranging between 1 and 5. Such findings are instrumental for industrial practitioners in formulating resource management strategies, particularly when stringent constraints are placed on ULs for RPs.

To offer a granular perspective, prediction errors for individual RPs are examined. A ridge plot presented in Fig. \ref{fig:4}(d) confirms that the error is consistently centred around zero across all RPs. Furthermore, a meticulous calibration analysis is performed through the assessment of empirical coverage probability across an array of prediction intervals. Fig. \ref{fig:4}(e) depicts a near-perfect alignment between the predicted probabilities and observed frequencies, substantiating the model's well-calibrated nature. This alignment underscores the model's adeptness in capturing the inherent uncertainties, suggesting a high degree of alignment between the posterior predictive distribution and the intrinsic probabilistic distribution governing the ULs.

\subsubsection{Real-World Evaluation of CLUSTER in Communications}\label{sec:Validation on realistic data}

We proceed to validate the effectiveness of CLUSTER using a dataset derived from real-world scenarios within the field of communications, predicting ULs of 302 BSs under various RA. In our endeavour to ensure privacy protection, we source this dataset from an anonymous city in China. The selected area is populated with an extensive network of 302 BSs, serving a dynamically changing population of approximately 7,000 users.

The deployment of the Na\"ive CLUSTER on the given dataset can provide insights into user clusters. During the inference stage, the pre-specified cluster count, \(W\), is gradually increased, with the inference procedure reiterated for multiple iterations. A subsequent application of the t-SNE technique to the MCMC-generated posterior samples for UPs facilitates dimensionality reduction. As depicted in Fig. \ref{fig:5}(a), a salient clustering topology manifests when \(W\) is configured at 40, thereby signalling the existence of approximately 40 latent clusters within the dataset.

Subsequently, the Complete CLUSTER is utilised for both inference and predictive tasks. The posterior distribution of the weights associated with the clusters in Fig. \ref{fig:5}(b) shows a rapid decrease as the number of clusters increases, which is consistent with the clustering nature of DPMM. With the truncation error threshold, as in Equation \eqref{eq:truncation_error}, trivially set at \(10^{-6}\) for demonstrative purposes, we obtain about 10 valid user clusters, as demonstrated in Fig. \ref{fig:6}(a). In practice, we can just retain the first 10 clusters with their corresponding weights and UPs, discarding the remaining clusters for computational efficiency. It serves as a testament to CLUSTER's remarkable capability to coalesce users sharing similar UP attributes into coherent and meaningful clusters.

It is worth noting that despite the apparent reduction in the number of clusters when transitioning from the Na\"ive to the Complete CLUSTER model, the performance metrics remain desirable. Fig. \ref{fig:6}(b) shows that the mean absolute error (MAE) for a majority of the BSs is minimal, an indicator of high general prediction accuracy. This observation gains further credibility from the clustering of nominal prediction errors around the zero mark, as illuminated in Fig. \ref{fig:6}(c).

The evaluation extends to the analysis of the standard deviations of these predictive outcomes, which are thoroughly presented in Fig. \ref{fig:6}(d) and \ref{fig:6}(e). To add another layer of validation, Fig. \ref{fig:6}(f) presents the calibration curve. Whilst not an exact overlay, the curve maintains a noteworthy alignment with the diagonal ideal line. This significant alignment serves to underline the superior capacity of the CLUSTER algorithm to faithfully represent the true distribution of ULs, especially in a setting characterised by its intricate and expansive scale.

\newpage
\section{Discussion}

In this study, we introduce and evaluate CLUSTER, a novel hierarchical Bayesian nonparametric model designed to address the challenge of predicting user transfer in the context of fluctuating RA. The demand for such a tool is paramount, as resource optimisation is crucial for sustaining economic viability and competitiveness across diverse fields.

A primary strength of CLUSTER lies in its capacity to offer enhanced interpretability and accuracy when confronted with intricate user transfer patterns. It fills a pressing gap in current modelling methodologies, many of which either lack predictive accuracy or struggle with an absence of interpretability. The latter is especially vital in industrial decision-making, where not only the outcome of a model is essential but also an understanding of how that result is achieved.

Through our model, we have shown how user clusters can be identified and utilised to bolster the modelling process. By harnessing these naturally arising clusters, CLUSTER can predict the user transfer at a macro level. This cluster-centred approach considers the similarities in UPs of different users, avoiding modelling UP for each individual user and thus driving computational efficiency.

The incorporation of the DPMM in our framework obviates the need for explicit pre-specification of the number of clusters, thus simplifying the modelling process and allowing for the adaptability that is absent in many traditional modelling techniques. This attribute endows CLUSTER with a versatility that enables it to adjust to a broad spectrum of scenarios and data structures, marking it a versatile instrument in the modelling toolkit.

Another notable accomplishment of our model is its proficiency in effectively quantifying prediction uncertainty. By presenting a posterior predictive distribution for the ULs associated with each RP, CLUSTER delivers a detailed representation of the potential outcomes based on the RA data. This intricate level of uncertainty quantification is seldom seen in other models, signifying a pivotal advancement that CLUSTER brings to the table. Trustworthy decision-making in industrial contexts often entails the comprehension and judicious management of uncertainty, making this facet of our framework especially influential.

Moreover, the compatibility of CLUSTER with data anonymisation protocols stands out, particularly in light of the growing emphasis on user privacy in data management and processing. This capability underscores the practical value of CLUSTER in real-world scenarios, where user data is plentiful but often needs to be anonymised to safeguard privacy.

Our study involves a series of experiments utilising both simulated and real-world datasets to assess CLUSTER. These experiments are pivotal in showcasing the merits of our approach. In real-world contexts, typified by intricate, large-scale resource management scenarios, CLUSTER excels in characterising the true underlying UL distribution by providing a posterior predictive distribution highly aligned with the observed data. These results underline the potential of CLUSTER in addressing industrial-scale challenges and stress its accuracy in prediction and uncertainty quantification.

Nevertheless, it is pertinent to recognise certain limitations of our strategy. Whilst CLUSTER's automated clustering mitigates the computational burden, managing extraordinarily large datasets or functioning within the confines of limited computational resources may still pose challenges. Furthermore, the choice of the PSF has a direct bearing on the ultimate prediction accuracy.

These observed limitations denote prospective focal points for future research and enhancement of our methodology. Efforts could be directed towards amplifying the computational efficiency of CLUSTER, possibly through the employment of more sophisticated machine learning methods or by leveraging distributed computing frameworks. Additionally, formulating strategies for the automatic determination of the PSF is important, as it would augment the model's flexibility.

In conclusion, our research introduces and validates a pioneering model for user transfer, setting the stage for further investigations in resource management strategies. Our findings have broad implications, offering the potential to enhance resource efficiency across diverse fields, including energy, transport, communications, etc. We look forward with keen interest to observing how CLUSTER will be integrated and adapted to spearhead efficient resource optimisation in these domains.

\begin{figure*}[t]  


\centering   \includegraphics[width=0.925\linewidth]{Figs/1.pdf}  \vspace{0.5cm} \caption{\linespread{1}\selectfont{}\textbf{User Transfer Overview with User Preferences and Resource Availability Influences.} In panel (\textbf{a}), the top figure illustrates how ULs for each RP are influenced by both the RP's RA, depicted by coloured spheres and UPs. Larger colour spheres specify higher RA, and proximity to RPs indicates a stronger preference. The bottom bar graphs provide an overall view of the user transfer amongst RPs. This study aims to learn from historical user transfer and to predict future transfer patterns under new RA configurations. Panel (\textbf{b}) analyses a representative simulated dataset composed of 3 RPs and 100 users. Here, the RA for each RP varies over time, leading to corresponding changes in ULs. This panel reveals a complex and stochastic correlation between the ULs and the RA (more details in \methods~and supplementary video). Panel (\textbf{c}) outlines the prediction workflow of CLUSTER (more details in Section \ref{sec:user_RA}). It shows that observed ULs are influenced by both the RA of each RP and the UP associated with each user, which are the inputs to the PSF. The output of this function is a vector on a standard simplex, interpreted as the probability of users connecting to each RP. The likelihood is then employed to generate the predicted ULs for each RP.
}
\label{fig:1}
\end{figure*}

\begin{figure*}[t]  


\centering   \includegraphics[width=0.925\linewidth]{Figs/3.pdf}  \vspace{0.5cm} \caption{\linespread{1}\selectfont{}\textbf{From t-SNE Visualisation to the Complete CLUSTER.} In panel (\textbf{a}), t-SNE visualises UPs generated by Na\"ive CLUSTER in a 2D plane. The pre-set cluster number \(W\) varies in separate tests, but approximately 10 clusters consistently emerge, demonstrating the robustness of our model. Panel (\textbf{b}) depicts the Complete CLUSTER, incorporating the DPMM. The stick-breaking process, key to the DPMM, is shown within a blue dashed line. Panel (\textbf{c}) presents a violin plot of cluster weights derived from MCMC-generated posterior samples. Some smaller weights are shown on a logarithmic scale for clarity. The first few clusters account for most of the weight, validating the truncation scheme. Whilst infinite clusters exist theoretically, only the first 10 weights are displayed for clarity. The orange dashed line denotes the truncation boundary by which only components with significant weights are retained.
}
\label{fig:3}
\end{figure*}

\begin{figure*}[t]  


\centering   \includegraphics[width=0.925\linewidth]{Figs/4.pdf}  \vspace{0.5cm} \caption{\linespread{1}\selectfont{}\textbf{Workflow and Performance Evaluation of Complete CLUSTER Using a Simulated Dataset. } In panel (\textbf{a}), Complete CLUSTER's workflow is depicted, comprising inference and prediction phases. The inference phase inputs historical RA-UL pairs to yield the posterior distribution of latent variables, of which only \(\bmrm{w}\) and \(\bmrm{L}\) are shown for clarity. New RA is introduced in the prediction phase to obtain the posterior predictive distribution of ULs, as per Equation \eqref{eq:post_pred}. The use of DPMM implies an infinite number of components in cluster weights and UPs. Performance is assessed using a simulated dataset. Panel (\textbf{b}) measures nominal prediction error across RPs and explanatory variables, with a histogram centred around 0, indicating generally accurate nominal predictions. Panel (\textbf{c}) shows prediction uncertainty for each RP and explanatory variables through standard deviation. Panel (\textbf{d}) offers a ridge plot of errors, highlighting a concentration around 0 despite minor variations amongst RPs, confirming accurate predictions. Panel (\textbf{e}) employs a calibration plot for uncertainty quantification. A green diagonal line represents ideal calibration; CLUSTER's line closely aligns, validating the alignment between the posterior predictive distribution and the true underlying distribution governing the ULs.
}
\label{fig:4}
\end{figure*}

\begin{figure*}[t]  


\centering   \includegraphics[width=0.925\linewidth]{Figs/5.pdf}  \vspace{0.5cm} \caption{\linespread{1}\selectfont{}\textbf{Insight into User Clustering with CLUSTER Using Real-world Dataset.} Panel (\textbf{a}) applies Na\"ive CLUSTER to real-world data. Increasing the pre-set number of user clusters, \(W\), yields a distinct clustering pattern at \(W = 40\) via 2D t-SNE visualisation, suggesting roughly 40 user clusters. In panel (\textbf{b}), the posterior distribution for weights using Complete CLUSTER is deployed. Posterior distributions of weights for each latent cluster, \(\bmrm{w}\), are log-scaled for clarity. Despite Na\"ive CLUSTER indicating around 40 user clusters, DPMM integration in Complete CLUSTER aggregates users with similar UPs, leading to fewer clusters with significant weights.
}
\label{fig:5}
\end{figure*}

\begin{figure*}[t]  


\centering   \includegraphics[width=0.925\linewidth]{Figs/6.pdf}  \vspace{0.5cm} \caption{\linespread{1}\selectfont{}\textbf{Analysing Complete CLUSTER Performance Metrics Using Real-world Base Station Dataset. } In panel (\textbf{a}), truncation error \(\varepsilon\paren{m}\) is plotted against the number of components, \(m\), on a log scale. A threshold at \(\delta = 10^{-6}\) reveals the first 10 components as sufficient. Panel (\textbf{b}) features a heatmap of the MAE for nominal prediction errors, each column representing a unique RP and rows indicating results for different explanatory variables, \(\bmrm{u}\paren{k}\). Panel (\textbf{c}) presents a histogram showing prediction errors across RPs and variables, displaying a focus around 0, confirming high general accuracy. Panels (\textbf{d}) and (\textbf{e}) outline the standard deviation of predictions for various RPs and explanatory variables. Panel (\textbf{f}) displays a calibration plot, used for uncertainty quantification on real-world data. The close alignment with the diagonal line corroborates CLUSTER's robust ability to accurately model the true UL distribution, even in complex, large-scale scenarios.}
\label{fig:6}
\end{figure*}

\newpage
\section{Methods}

\subsection{Simulated Dataset Generation for CLUSTER Validation}

We perform the validation of CLUSTER by generating a simulated dataset, simulating the interactions between BSs and mobile phone users. Herein, we set \(N\) and \(M\) as the number of BSs and users, respectively. To determine the positions of the BSs, \(\paren{\upphi_{i}^{\mathrm{x}}, \upphi_{i}^{\mathrm{y}}}\), we employ a uniform distribution within a unit area, as formalised in the equation below:
\begin{linenomath*}
\begin{equation}\label{eq:simulated_BS_loc}
    \paren{\upphi_{i}^{\mathrm{x}}, \upphi_{i}^{\mathrm{y}}} \sim \operatorname{Uniform}\paren{\brackets{0, 1} \times \brackets{0, 1}}, i = 1, \ldots, N.
\end{equation}
\end{linenomath*}
Next, to simulate users, we randomly select \(M\) points within the same region to serve as the users' `homes'. This selection process can be represented by the following equation:
\begin{linenomath*}
\begin{equation}\label{eq:simulated_user_loc}
    \paren{\upmu_{j}^{\mathrm{x}}, \upmu_{j}^{\mathrm{y}}} \sim \operatorname{Uniform}\paren{\brackets{0, 1} \times \brackets{0, 1}}, j = 1, \ldots, M,
\end{equation}
\end{linenomath*}
where \(\paren{\upmu_{j}^{\mathrm{x}}, \upmu_{j}^{\mathrm{y}}}\) denotes the locations of users' homes.

To mimic the mobility of users, we incorporate a mean-reverting pattern around these homes. Each user is allotted individual `mobility' and `reverting rate' parameters, denoted by \(\upsigma\) and \(\uptheta\), respectively. Here, \(\upsigma \sim \operatorname{HalfNormal}\paren{K}\) serves as the mobility parameter, and \(\uptheta \sim \operatorname{HalfNormal}\paren{r}\) represents the reverting rate. \(K\) and \(r\) are preset parameters. These parameters are integrated into the Ornstein–Uhlenbeck process to formulate user movement:
\begin{linenomath*}
\begin{equation}\label{eq:simulated_OU_continuous}
\begin{aligned}
    \mathrm{d}\mathrm{p}^{\mathrm{x}}_{j}\paren{t} &= - \uptheta_{j} \paren{\mathrm{p}^{\mathrm{x}}_{j}\paren{t} - \upmu_{j}^{\mathrm{x}}} \mathrm{d}t + \upsigma_{j} \mathrm{d}\mathrm{W}\paren{t},  \\
    \mathrm{d}\mathrm{p}^{\mathrm{y}}_{j}\paren{t} &= - \uptheta_{j} \paren{\mathrm{p}^{\mathrm{y}}_{j}\paren{t} - \upmu_{j}^{\mathrm{y}}} \mathrm{d}t + \upsigma_{j} \mathrm{d}\mathrm{W}\paren{t},
\end{aligned}
\end{equation}
\end{linenomath*}
where \(\paren{\mathrm{p}^{\mathrm{x}}_{j}\paren{t}, \mathrm{p}^{\mathrm{y}}_{j}\paren{t}}\) is the real-time location of the \(j\)-th user, and \(\mathrm{W}\paren{t}\) signifies the Wiener process in Equation \eqref{eq:simulated_OU_continuous}. 

For computational simulations, we convert the continuous-time differential equation into a discrete form by leveraging the Euler-Maruyama method:
\begin{linenomath*}
\begin{equation}\label{eq:simulated_OU_discrete}
\begin{aligned}
    \mathrm{p}^{\mathrm{x}}_{j}\paren{k + 1} &= \upmu_{j}^{\mathrm{x}} + \paren{\mathrm{p}^{\mathrm{x}}_{j}\paren{k} - \upmu_{j}^{\mathrm{x}}} e^{-\uptheta \Delta t} + \sqrt{\frac{\upsigma_{j}^{2}}{2 \uptheta_{j}}\paren{1 - e^{-2\uptheta_{j} \Delta t}}} \upepsilon\paren{k},  \\
    \mathrm{p}^{\mathrm{y}}_{j}\paren{k + 1} &= \upmu_{j}^{\mathrm{y}} + \paren{\mathrm{p}^{\mathrm{y}}_{j}\paren{k} - \upmu_{j}^{\mathrm{y}}} e^{-\uptheta \Delta t} + \sqrt{\frac{\upsigma_{j}^{2}}{2 \uptheta_{j}}\paren{1 - e^{-2\uptheta_{j} \Delta t}}} \upepsilon\paren{k}.
\end{aligned}
\end{equation}
\end{linenomath*}
In these equations, \(\upepsilon \paren{k} \sim \mathcal{N}\paren{0, 1}\), and \(\Delta t\) denotes the sampling period. 

The simulation process spans a total of \(S\) timesteps, where \(S\) is a positive integer. During this period, the operational status of each BS represented as \(\mathrm{u}_{i}\paren{k}\), can toggle between an `on' or `off' state, signifying a binary mode. This operational status can be expressed as:
\begin{linenomath*}
\begin{equation}\label{eq:simulated_u_dist}
    \mathrm{u}_{i}\paren{k} \sim \operatorname{Bernoulli}\paren{p},
\end{equation}
\end{linenomath*}
where \(p\) is a predefined probability, and \(p \in \paren{0, 1}\).

This binary state changes every \(d\) sampling periods. To maintain seamless user connectivity, we always ensure that at least one BS is operational at any given time. We assign \(\mathrm{c}_{j}\paren{k}\) as the symbol denoting a user's connection choice. In our simulation, this choice is designed to be the nearest BS, represented by the following equation:
\begin{linenomath*}
\begin{equation}\label{eq:simulated_c_choice}
    \mathrm{c}_{j}\paren{k} = \argmin{i}{\sqrt{\paren{\mathrm{p}^{\mathrm{x}}_{j}\paren{k} - \upphi^{\mathrm{x}}_{i}}^2 + \paren{\mathrm{p}^{\mathrm{y}}_{j}\paren{k} - \upphi^{\mathrm{y}}_{i}}^2}}.
\end{equation}
\end{linenomath*}

We define the UL for the \(i\)-th BS during recording period \(n\), \(\mathrm{x}_{i}\paren{n}\), as the arithmetic mean of the number of users connected over \(d\) sampling periods. We quantify the size of the simulated dataset, \(D\), by \(D = \frac{S}{d}\). The parameters \(S\) and \(d\) are appropriately chosen to ensure \(D\) is an integer. For each BS, we compute the valid UL during each recording period \(n\), as follows:
\begin{linenomath*}
\begin{equation}\label{eq:simulated_x_define}
    \mathrm{x}_{i}\paren{n} = \frac{1}{d} \sum_{k = \paren{n - 1}d + 1}^{nd} \sum_{j=1}^{M} \bm{\mathrm{1}}_{i} \paren{\mathrm{c}_{j}\paren{k}}, n = 1, \ldots, D,
\end{equation}
\end{linenomath*}
where \(\map{\bm{\mathrm{1}}_{i}}{\until{N}}{\braces{0, 1}}\) serves as an indicator function, which is explicitly defined by the following relation:
\begin{linenomath*}
\begin{equation}\label{eq:simulated_indicator_func}
    \bm{\mathrm{1}}_{i} \paren{\mathrm{c}_{j}\paren{k}} =
    \begin{cases} 
        1, & \text{if } \mathrm{c}_{j}\paren{k} = i , \\
        0, & \text{if } \mathrm{c}_{j}\paren{k} \neq i .
    \end{cases}
\end{equation}
\end{linenomath*}
The RA of the BSs during a given recording period is defined as the proportion of time when each BS is set to `on'. Mathematically, it can be expressed as:
\begin{linenomath*}
\begin{equation}\label{eq:simulated_u_define}
    \mathrm{u}_{i}\paren{n} = \frac{1}{d} \sum_{k = \paren{n - 1}d + 1}^{nd} \mathrm{u}_{i}\paren{k}, n = 1, \ldots, D.
\end{equation}
\end{linenomath*}

\subsection{Ground-Truth Validation with the Real-world Base Station Dataset}

In order to provide empirical validation for CLUSTER, we utilise a real-world BS dataset sourced from an anonymised city in China. The city's name is withheld to maintain privacy and confidentiality.

The dataset contains information spanning over a period of 82 days. During this time, both the RA of the BSs and the ULs are reported. Throughout this period, the average user count remains around 7,000. This figure, however, is an aggregated average; real-time user counts can fluctuate significantly due to the influence of the diurnal cycle.

The focus of our study is on a specified zone within the city that encompasses 302 BSs. This zone is notably self-contained, allowing for a practical approximation that the users primarily transfer within these 302 BSs. Thus, the overall quantity of ULs remains invariant upon RA variations.

The ULs within this dataset are quantified every 15 minutes. The metric for these loads is the average number of users connected to each BS during the 15-minute interval. The RA of BSs is determined by the proportion of each 15-minute interval during which the BS is switched on. This is analogous to the simulated dataset used in our model.

\subsection{Formulation of CLUSTER for Predicting User Transfer}

The primary aim of this research is to model the prediction of user transfer in light of changes in the RA of RPs. Accordingly, the design of CLUSTER is expected to leverage the complex interplay between the RA at the \(k\)-th time step, \(\bmrm{u}\paren{k}\), and the subsequent ULs, \(\bmrm{x}\paren{k + 1}\). We begin with Na\"ive CLUSTER, the structural essence of which is elucidated by the following equations:
\begin{linenomath*}
\begin{equation}\label{eq:naive_CLUSTER}
\begin{aligned}
    \frac{\bmrm{x}\paren{k + 1}}{M} &\sim \operatorname{Dirichlet}\paren{\mathrm{c}\bmrm{p}\paren{k + 1}}  \\
    \mathrm{c} &\sim \operatorname{Gamma}\paren{a, b} \\
    \bmrm{p}\paren{k + 1} &= \paren{\bmrm{A}\paren{k + 1}}^{\top} \bmrm{w}  \\
    \bmrm{A}\paren{k + 1} &= \operatorname{diag}\paren{\hat{\bmrm{A}}\paren{k + 1}\bmrm{1}_{N}}^{-1}\hat{\bmrm{A}}\paren{k + 1}  \\
    \hat{\bmrm{A}}\paren{k + 1} &= \bmrm{L}  \operatorname{diag}\paren{\bmrm{u}\paren{k}}  \\
    \bmrm{L} &= \brackets{\bmrm{l}^{1}, \ldots, \bmrm{l}^{W}}^{\top}  \\
    \bmrm{l}^{1}, \ldots, \bmrm{l}^{W} &\sim \operatorname{Dirichlet}\paren{\bmrm{1}_{N}}  \\
    \bmrm{w} &\sim \operatorname{Dirichlet}\paren{\bmrm{1}_{W}}.
\end{aligned}
\end{equation}
\end{linenomath*}
Herein, \(M\), representing the total quantity of ULs, is assumed known in advance. Selected hyperparameters \(a\) and \(b\) make the \(\operatorname{Gamma}\paren{a, b}\) a weakly informative prior. The \(W\), manually pre-set, signifies the number of user clusters. The vector \(\bmrm{w}\) depicts the weight apportioned to each cluster, where \(\bmrm{w} \in \Delta^{W - 1}\). A detailed derivation of these equations is elaborated upon in the Supplementary Information. 

The Complete CLUSTER, integrating the DPMM, is also explored in our study, where a stick-breaking process is employed to construct the Dirichlet process. The symbol \(W\) denotes the count of significant clusters whilst \(\bmrm{w}\) denotes the weight of each cluster, adhering to the domain \(\bmrm{w} \in \Delta^{W - 1}\). The comprehensive model of the Complete CLUSTER is expressed as:
\begin{linenomath*}
\begin{equation}\label{eq:full_CLUSTER}
\begin{aligned}
    \frac{\bmrm{x}\paren{k + 1}}{M} &\sim \operatorname{Dirichlet}\paren{\mathrm{c}\bmrm{p}\paren{k + 1}}  \\
    \mathrm{c} &\sim \operatorname{Gamma}\paren{a, b} \\
    \bmrm{p}\paren{k + 1} &= \paren{\bmrm{A}\paren{k + 1}}^{\top} \bmrm{w}  \\
    \bmrm{A}\paren{k + 1} &= \operatorname{diag}\paren{\hat{\bmrm{A}}\paren{k + 1}\bmrm{1}_{N}}^{-1}\hat{\bmrm{A}}\paren{k + 1}  \\
    \hat{\bmrm{A}}\paren{k + 1} &= \bmrm{L} \operatorname{diag}\paren{\bmrm{u}\paren{k}}  \\
    \bmrm{L} &= \brackets{\bmrm{l}^{1}, \ldots, \bmrm{l}^{W}}^{\top}  \\
    \bmrm{l}^{1}, \ldots, \bmrm{l}^{W} &\sim \operatorname{Dirichlet}\paren{\bmrm{1}_{N}}  \\
    \bmrm{w} &= \brackets{\mathrm{w}_{1}, \ldots, \mathrm{w}_{W}}^{\top}  \\
    \mathrm{w}_{j} &= \upbeta_{j} \prod_{k = j-1}^{j} \paren{1 - \upbeta_{k}}  \\
    \upbeta_{1}, \ldots, \upbeta_{W} &\sim \operatorname{Beta}\paren{1, \upalpha}  \\
    \upalpha &\sim \operatorname{Gamma}\paren{c, d}.
\end{aligned}
\end{equation}
\end{linenomath*}
In the selection of priors, we opt for distributions that are either non-informative or weakly informative. Specifically, \(\operatorname{Dirichlet}\paren{\bmrm{1}_{N}}\) and \(\operatorname{Gamma}\paren{c, d}\) are chosen to reflect the lack of prior knowledge concerning UPs and the concentration parameter respectively.

Whilst some sampling techniques for Dirichlet processes allow for dynamic expansion of the cluster number, for simplicity, our approach employs truncation upon reaching \(W\) clusters that satisfy the following truncation error criteria. The truncation error, \(\varepsilon\paren{m}\), is defined as per Equation \eqref{eq:truncation_error},
\begin{linenomath*}
\begin{equation}\label{eq:truncation_error}
\varepsilon\paren{m} = 1 - \sum_{k = 1}^{m}\mathrm{w}_{k} .
\end{equation}
\end{linenomath*}
The number of retained components is based on the threshold imposed upon the cumulative mixture weights:
\begin{linenomath*}
\begin{equation}\label{eq:W_truncate}
W = \argmin{m}{\paren{\varepsilon\paren{m} \leq \delta}} ,
\end{equation}
\end{linenomath*}
where \(\delta\) represents a pre-defined threshold.

\subsection{Posterior Predictive Checks using MCMC Approximations}\label{met:amalgamation}

In the ensuing analysis, we delve into the mathematical underpinnings that assess the quality of our predictive model, with particular attention to prediction error and uncertainty quantification. First, we discuss the performance metrics from a mathematical standpoint. The mathematical expectation of the predicted ULs for each RP, denoted as \(\tilde{\bmrm{x}}\paren{k + 1}\), is evaluated to serve as the nominal prediction result, as in Equation \eqref{eq:mean_pred}:
\begin{linenomath*}
\begin{equation}\label{eq:mean_pred}
    \tilde{\bmrm{x}}\paren{k + 1} = \int_{\bmrm{\chi}} \int_{\bm{\Theta}} \bmrm{x}\paren{k + 1} p\paren{\bmrm{x}\paren{k + 1} | \bmrm{x}\paren{k}, \bmrm{u}\paren{k}, \bm{\uptheta}, \mathcal{M}} p\paren{\bm{\uptheta} | \mathcal{D}, \mathcal{M}} \mathrm{d} \bm{\uptheta} \mathrm{d} \bmrm{x}\paren{k + 1},
\end{equation}
\end{linenomath*}
where \(\bmrm{\chi}\) represents the variable space satisfying \(\norm{\bmrm{x}\paren{k}}_{1} = \norm{\bmrm{x}\paren{k + 1}}_{1}\). We also analyse the nominal prediction error \(\bmrm{e}\left(k + 1\right)\):
\begin{linenomath*}
\begin{equation}\label{eq:error_pred}
    \bmrm{e}\paren{k + 1} = \tilde{\bmrm{x}}\paren{k + 1} - \hat{\bmrm{x}}\paren{k + 1},
\end{equation}
\end{linenomath*}
with \(\hat{\bmrm{x}}\left(k + 1\right)\) representing the groundtruth value. To evaluate the degree of uncertainty in predictions, we utilise the SD of the prediction results for each RP. As in Equation \eqref{eq:error_std}, the SD for the \(i\)-th RP is formulated as follows:
\begin{linenomath*}
\begin{multline}\label{eq:error_std}
    \sigma \paren{\mathrm{x}_i} = \\
    \sqrt{\int_{\bmrm{\chi}_{i}} \int_{\bm{\Theta}} \paren{\mathrm{x}_{i}\paren{k + 1} - \tilde{\mathrm{x}}_{i}\paren{k + 1}}^{2} p\paren{\mathrm{x}_{i}\paren{k + 1} | \bmrm{x}\paren{k}, \bmrm{u}\paren{k}, \bm{\uptheta}, \mathcal{M}} p\paren{\bm{\uptheta} | \mathcal{D}, \mathcal{M}} \mathrm{d}\mathrm{\bm{\uptheta}}\mathrm{d}\mathrm{x}_{i}\paren{k + 1}}  ,
\end{multline}
\end{linenomath*}
where \(\bmrm{\chi}_{i}\) denotes the subspace in \(\bmrm{\chi}\) for the \(i\)-th RP. 

However, the direct evaluation of Equation \eqref{eq:mean_pred} and Equation \eqref{eq:error_std} is problematic owing to the paucity of closed-form solutions. To circumvent this, we employ an approximation of these statistics using the MCMC-generated samples from the posterior predictive distribution. This approach obviates the need to evaluate the high-dimensional integral.

We consider that posterior predictive checks use \( \bmrm{u}\paren{k} \) with \( R \) distinct values in the dataset, while \( \bmrm{x}\paren{k} \) is assumed to have \( N \) components. We subsequently define \(\mathrm{x}_{i}^{r}\paren{k + 1}\) as the response value of the \(i\)-th RP for the \(r\)-th \(\bmrm{u}\paren{k}\), where \(i \in \until{N}\) and \(r \in \until{R}\). The \(\mathrm{x}_{i}^{r,q}\paren{k + 1}\) denotes the \(q\)-th MCMC-generated sample for \(\mathrm{x}_{i}^{r}\paren{k + 1}\) from the posterior predictive distribution, where \(q \in \until{Q}\). Building on this, the nominal prediction result and the standard deviations for each response variable can be computed in line with the following equations:
\begin{linenomath*}
\begin{equation}\label{eq:real_mean_pred}
    \tilde{\mathrm{x}}_{i}^{r}\paren{k + 1} \approx \frac{1}{Q} \sum_{q=1}^{Q} \mathrm{x}_{i}^{r, q}\paren{k + 1},
\end{equation}
\end{linenomath*}
\begin{linenomath*}
\begin{equation}\label{eq:real_pred_std}
    \sigma\paren{\mathrm{x}_{i}^{r}\paren{k + 1}} \approx \sqrt{\frac{1}{Q - 1} \sum_{q=1}^{Q} \paren{\mathrm{x}_{i}^{r, q}\paren{k + 1} - \tilde{\mathrm{x}}_{i}^{r}\paren{k + 1}}^{2}}.
\end{equation}
\end{linenomath*}

The multivariate nature of \(\bmrm{x}\paren{k}\) introduces complexities when attempting to systematically gauge prediction error and conduct uncertainty quantification. However, recognising that separate components of \(\bmrm{x}\paren{k}\) exhibit common traits and are on comparable scales, it is reasonable to combine the results for different \(\bmrm{u}\paren{k}\)'s across varied RPs. Whilst this approximation, as visualised in histograms in Fig. \ref{fig:4} and Fig. \ref{fig:6}, might not be fully rigorous, it does offer insights into the efficacy of CLUSTER across diverse scenarios using a standardised benchmark.

\subsection{Calibration Evaluation of CLUSTER with Multivariate Responses}\label{sec:reliability_plot}

The assessment of model calibration, as delineated in \cite{gelman2013bayesian}, stipulates that posterior means ought to be accurate on average. The \(P\%\) intervals should theoretically encompass the true values \(P\%\) of the time for different \(P\), where \(P \in \brackets{0, 100}\). Specifically, for a specified \(P\) and a particular \(\bmrm{u}\paren{k}\), the \(P\%\) highest density interval (HDI) of the posterior predictive distribution is ascertained for each response variable, \(\mathrm{x}_{i}^{r}\paren{k + 1}\). Herein, \(P\) is denominated as the `nominal coverage probability'. Then, we determine the fraction of the true response value encapsulated within the \(P\%\) HDI across various \(\bmrm{u}\paren{k}\)'s and RPs, termed as the `empirical coverage probability'. 

The assessment of calibration becomes intricate owing to the multivariate character of \(\bmrm{x}\paren{k + 1}\). Compounding this complexity is the challenge of repeating the same experiment numerous times with \(\bmrm{u}\paren{k}\) and \(\bmrm{x}\paren{k}\) held constant, which makes the evaluation of the empirical coverage probability difficult. By recognising the similarities in characteristics and scales for each component of \(\bmrm{x}\paren{k + 1}\), as well as the comparable variations in \(\bmrm{u}\paren{k}\) across the dataset, we adopt an approximation strategy. Aggregating the prediction results for different \(\bmrm{u}\paren{k}\)'s across various RPs, and treating them as repeated experiments, allows for a systematic appraisal of CLUSTER's calibration performance, offering a comprehensive insight into the model's calibration. Suppose that \(\mathbb{P}_{i}^{r, P} = \brackets{\mathrm{x}_{i}^{r, P^\mathrm{lower}}, \mathrm{x}_{i}^{r, P^\mathrm{upper}}}\) is the \(P\%\) HDI for the \(i\)-th RP and \(r\)-th \(\bmrm{u}\paren{k}\) in the dataset, then the overall empirical coverage probability \(\tilde{P}\%\) can be given as Equation \eqref{eq:empirical_coverage_prob}:
\begin{linenomath*}
\begin{equation}\label{eq:empirical_coverage_prob}
    \tilde{P} = \frac{\sum_{i=1}^{N} \sum_{r=1}^{R} \bm{\mathrm{1}}_{\mathbb{P}_{i}^{r, P}}\paren{\hat{\mathrm{x}}_{i}^{r}}}{NR} \times 100\% ,
\end{equation}
\end{linenomath*}
where \(\hat{\mathrm{x}}_{i}^{r}\) is the groundtruth value in the dataset, and the indicator function \(\bm{\mathrm{1}}_{\mathbb{P}_{i}^{r, P}}\paren{\hat{\mathrm{x}}_{i}^{r}}\) is defined as
\begin{linenomath*}
\begin{equation}\label{eq:empirical_coverage_indicator}
    \bm{\mathrm{1}}_{\mathbb{P}_{i}^{r, P}}\paren{\hat{\mathrm{x}}_{i}^{r}} =  
    \begin{cases} 
        1, & \text{if } \hat{\mathrm{x}}_{i}^{r} \in \mathbb{P}_{i}^{r, P} , \\
        0, & \text{if } \hat{\mathrm{x}}_{i}^{r} \notin \mathbb{P}_{i}^{r, P} .
    \end{cases}
\end{equation}
\end{linenomath*}

This methodology is subsequently repeated for multiple \(P\) values and is depicted in the reliability plot in Fig. \ref{fig:4}(e) and Fig. \ref{fig:6}(f). In a perfectly calibrated model, the plotted markers would coincide precisely with the diagonal line. The disparity between these markers and the diagonal line offers a quantitative measure for our model's calibration evaluation. A model with smaller deviations is better calibrated, as it more accurately reflects the inherent uncertainties in the data.

In the case of the real-world dataset, a further step of post-processing the posterior predictive samples proves advantageous. This involves shrinking the MCMC-generated posterior predictive samples towards their mean, a method aimed at offsetting the undue increase in the distribution's variance caused by outliers of ULs in the training set. In both the training and testing sets, the outliers exhibit a uniform pattern, distinguished by a specific degree of expansion in the variance of the posterior predictive distribution. By judiciously reducing this variance while preserving the mean, the posterior predictive distribution may be calibrated, thereby reinstating the effectiveness of uncertainty quantification. This process can be expressed mathematically as:
\begin{linenomath*}
\begin{equation}\label{eq:posterior_predictive_sample_post_process}
    \bar{\mathrm{x}}_{i}^{r, q} = \frac{\mathrm{x}_{i}^{r, q} - \tilde{\mathrm{x}}_{i}^{r}}{s} + \tilde{\mathrm{x}}_{i}^{r},
\end{equation}
\end{linenomath*}
where \(\bar{\mathrm{x}}_{i}^{r, q}\) represents the shrunk sample, and \(s\) denotes the shrinkage parameter. The optimal value for \(s\) can be ascertained through cross-validation.

\subsection{Computational Details} 

The MCMC algorithm was used for latent variable inference, with PyMC as the chosen computational framework. All analyses were carried out in Python 3.11, with several open-source Python packages (Pandas, NumPy, SciPy, scikit-learn, Matplotlib, seaborn, Xarray, JAX, and PyMC) utilised. The computational process was executed on an infrastructure equipped with 8 NVIDIA RTX 4090 GPUs, collectively supported by a 2.8 GHz AMD EPYC7573 processor and 500 GB of RAM. This configuration facilitated the completion of the computation within approximately 8 hours.

\subsection{Data Availability}

The simulated datasets can be generated from the code in the repository provided in the Code Availability section. The simulated datasets that support the findings of this study are available at https://github.com/dongxu-lei/CLUSTER. The real-world base station data used in this study were collected in cooperation with China United Network Communications Group Co., Ltd. in a de-identified format. The de-identified data are available via email: \href{mailto:21B904013@stu.hit.edu.cn}{21B904013@stu.hit.edu.cn} for academic research purposes only. All other data in this study are available from the corresponding author upon reasonable request.

\subsection{Code Availability} 

The source code for simulated dataset generation, the inference and prediction scripts for CLUSTER, and the simulated dataset used to support the findings of this study can be found at https://github.com/dongxu-lei/CLUSTER.

\section*{Acknowledgements}

This work was supported in part by the Joint Funds of the National Natural Science Foundation of China (U20A20188), in part by the National Natural Science Foundation of China (62303403, 62303402), and in part by XPLORER PRIZE.

\section*{Author contributions}

S.Z. and H.G. conceived the study. D.L. undertook the theoretical analysis, executed numerical studies, and wrote the manuscript and Supplementary Information. D.L. and X.L. were responsible for figure preparation. X. L. and X.Y. aided in data analysis. X. L., Z.L. and W.S. contributed to software development. J.Q. proofread the manuscript. All authors reviewed and approved the final manuscript.

\section*{Competing interests}
The authors declare no competing interests.

\bibliography{ref}

 \nolinenumbers
 
\printfigures

\end{document}



\bibliographystyle{naturemag}
\title{Supplement to ``Characterising User Transfer Amid Industrial Resource Variation: A Bayesian Nonparametric Approach''}

\author{Dongxu Lei}
\affiliation{Research Institute of Intelligent Control and Systems, Harbin Institute of Technology, Harbin, China}

\author{Xiaotian Lin}
\affiliation{Intelligent Control and Systems Research Center, Yongjiang Laboratory, Ningbo, China }

\author{Xinghu Yu}
\affiliation{Ningbo Institute of Intelligent Equipment Technology Company, Ltd., Ningbo, China}

\author{Zhan Li}
\affiliation{Research Institute of Intelligent Control and Systems, Harbin Institute of Technology, Harbin, China }

\author{Weichao Sun}
\affiliation{Research Institute of Intelligent Control and Systems, Harbin Institute of Technology, Harbin, China  }

\author{Jianbin Qiu}
\affiliation{Research Institute of Intelligent Control and Systems, Harbin Institute of Technology, Harbin, China   }

\author{Songlin Zhuang}
\affiliation{Intelligent Control and Systems Research Center, Yongjiang Laboratory, Ningbo, China}

\author{Huijun Gao}
\affiliation{Research Institute of Intelligent Control and Systems, Harbin Institution of Technology, Harbin, China    \\
Intelligent Control and Systems Research Center, Yongjiang Laboratory, Ningbo, China\\ To whom correspondence should be addressed: songlin-zhuang@ylab.ac.cn, hjgao@hit.edu.cn
}

\date{\today}

\maketitle

\tableofcontents

\newpage

\section{List of Abbreviations}

In the interest of clarity and conciseness, the following abbreviations are employed throughout this paper:

\begin{itemize}
    \item RA: Resource Availability
    \item RP: Resource Provider
    \item UL: User Load
    \item UP: User Preference
    \item BS: Base Station
    \item CLUSTER: Characterising Latent User Structure Through Evidence Refinement
    \item DPMM: Dirichlet Process Mixture Model
    \item MAE: Mean Absolute Error
    \item PS: Preference Score
    \item PSF: Preference Score Function
    \item MCMC: Markov Chain Monte Carlo
    \item t-SNE: t-Distributed Stochastic Neighbour Embedding
    \item SD: Standard Deviation
    \item HDI: Highest Density Interval
    \item CI: Credible Interval
\end{itemize}

\newpage
\section{Facilitating Variance Modelling through Dirichlet Distribution}\label{sec:multinomian_deprecated}

\begin{figure*}[t]  

\centering   \includegraphics[width=0.925\linewidth]{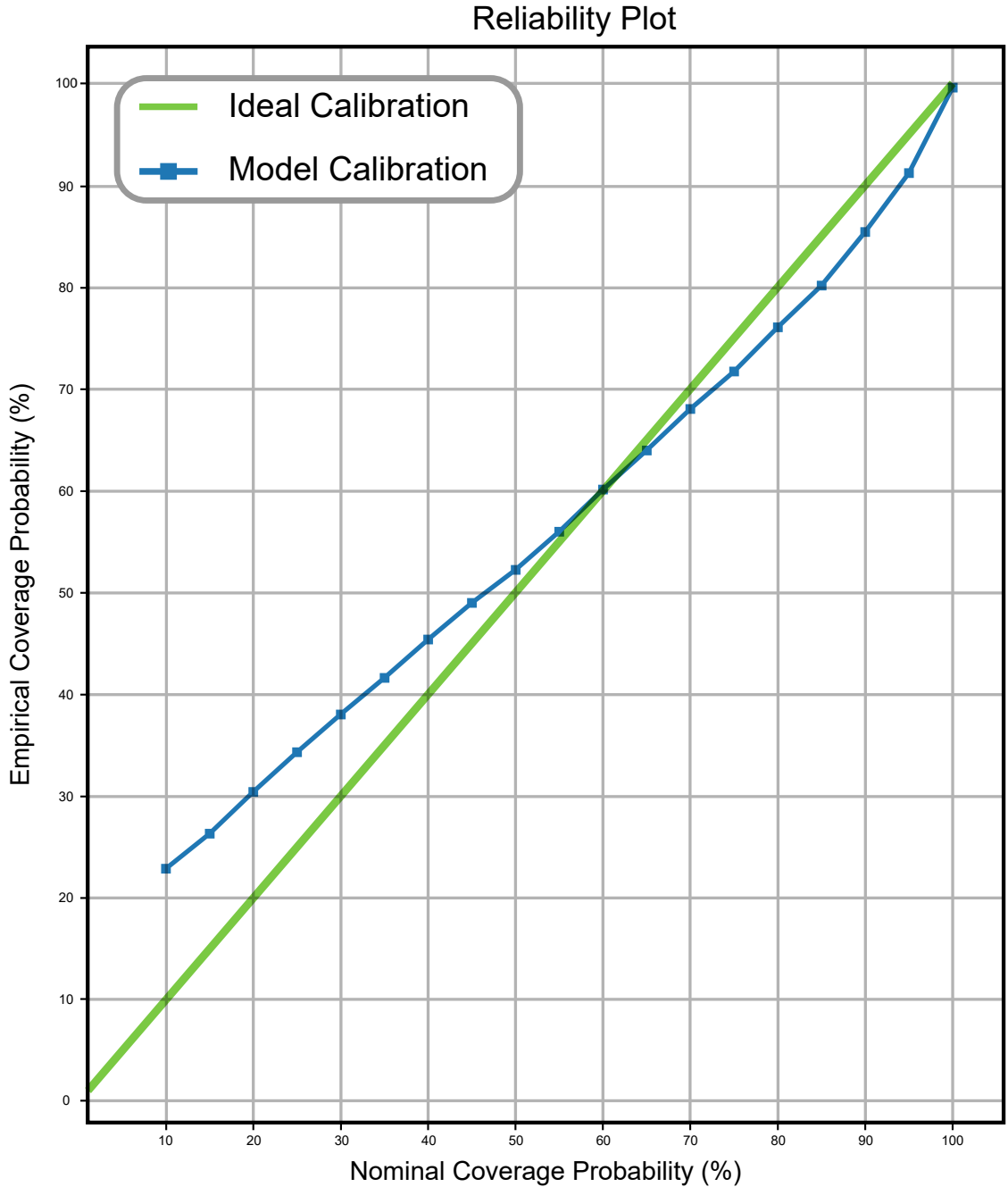}  \vspace{0.5cm} \caption{\linespread{1}\selectfont{}\textbf{Reliability plot assessing uncertainty quantification with the multinomial likelihood in the simulated dataset.} Reliability plot evaluates uncertainty quantification using a simulated dataset and multinomial distribution as the likelihood for CLUSTER. Notable deviation of the calibration curve from the diagonal line reveals suboptimal performance in uncertainty quantification. The inconsistent positioning of the curve relative to the diagonal complicates straightforward variance-based calibration.
}
\label{fig:Synthetic_DPMM_reliability_nomial}
\end{figure*}

In the final stages of modelling the likelihood, we must designate a distribution for sampling ULs. Although the multinomial distribution often serves as a standard selection for count data, its use is hindered by two critical limitations specific to our case. Firstly, the nature of certain aggregate data, such as average data, demands that ULs be considered continuous values. This is in contrast to the discrete support of the multinomial distribution, rendering the two incompatible. Secondly, the multinomial distribution stipulates that the variance of the proportion of ULs associated with each RP is constrained by the proportion vector, \(\bmrm{p}\paren{k + 1}\), and the total amount of ULs, \(M\), as outlined in Equation \eqref{eq:multi_var}
\begin{linenomath*}
\begin{equation}\label{eq:multi_var}
\operatorname{Var}\paren{\frac{\mathrm{x}_{i}\paren{k + 1}}{M}} = \frac{\mathrm{p}_{i}\paren{k + 1}\paren{1 - \mathrm{p}_i\paren{k + 1}}}{M}, i = 1, \ldots, N.
\end{equation}
\end{linenomath*}
This lack of flexibility of variance modelling compromises the uncertainty quantification capacity. Evidence of this shortfall is further elaborated in Fig. \ref{fig:Synthetic_DPMM_reliability_nomial}. In the model that employs the multinomial distribution, the calibration curve's noticeable deviation from the ideal diagonal line serves as empirical evidence of the model's shortcomings in calibration. As the nominal coverage probability ascends, the curve exhibits inconsistent behaviour: initially lying above the diagonal line and subsequently transitioning below it. This lack of consistency complicates any attempts to rectify the situation using the variance-based calibration approach delineated in the main text. Specifically, the erratic nature of the curve prevents a single adjustment—either inflation or deflation—of the posterior predictive distribution from being sufficient to align both segments of the curve closely with the diagonal line.

These shortcomings underscore the need for an alternative probabilistic distribution with continuous support that can independently model the variance of the UL proportions. Our solution employs the Dirichlet distribution, complemented by a positive latent concentration variable \(\mathrm{c}\), collectively capturing the mean proportion of ULs and the variance, as in Equation \eqref{eq:x_Dirichlet}.

This design choice permits a logical extension to prove that the mean and variance of the UL proportions can be formalised as expressed in Equation \eqref{eq:dirichlet_likelihood_mean} and Equation \eqref{eq:dirichlet_likelihood_var}:
\begin{linenomath*}
\begin{equation}\label{eq:dirichlet_likelihood_mean}
\mathbb{E}\paren{\frac{\bmrm{x}\paren{k + 1}}{M}} = \bmrm{p}\paren{k + 1},
\end{equation}
\end{linenomath*}
\begin{linenomath*}
\begin{equation}\label{eq:dirichlet_likelihood_var}
\operatorname{Var}\paren{\frac{\mathrm{x}_{i}\paren{k + 1}}{M}} = \frac{\mathrm{p}_{i}\paren{k + 1}\paren{1 - \mathrm{p}_{i}\paren{k + 1}}}{\mathrm{c} + 1}.
\end{equation}
\end{linenomath*}
Hence, this choice decouples the total quantity of ULs, \(M\), from the variance, allowing for independent modelling through the concentration variable \(\mathrm{c}\). This preserves the mathematical expectation of UL proportions, leading to a more adaptable model representation.

\newpage
\section{Structure and Specification of the Na\"ive CLUSTER Model}\label{sec:naive_CLUSTER}

In this section, we provide an overview of the rationale behind the structure of the Na\"ive CLUSTER model. One of its notable features is the manual specification of the number of user clusters by the model designer, typically based on expert knowledge or ad-hoc considerations.

The first step in the Na\"ive CLUSTER approach is to assign weights to each user cluster. Since all the clusters collectively represent the entire ULs considered in our problem, the sum of the weights assigned to each user cluster must equal exactly 1, as indicated in Equation \eqref{eq:w_norm_1}
\begin{linenomath*}
\begin{equation}\label{eq:w_norm_1}
    \norm{\bmrm{w}}_{1} = 1,
\end{equation}
\end{linenomath*}
where \(\norm{\cdot}_{1}\) denotes the \(\ell^{1}\) norm. It is important to note that, in general, we do not have prior knowledge or distinguishing characteristics of individual user clusters. The parameters associated with each cluster exhibit exchangeability. Therefore, to account for our lack of prior information about the individual clusters, we aim to employ a noninformative prior for the distribution of cluster weights, denoted as \(\bmrm{w}\). Mathematically, for each component \(\mathrm{w}_{j}\) of \(\bmrm{w}\), where \(j \in \until{W}\) and \(W\) represents the predetermined number of clusters, Equation \eqref{eq:w_0_1} should be satisfied
\begin{linenomath*}
\begin{equation}\label{eq:w_0_1}
    \mathrm{w}_{j} \in \brackets{0, 1}.
\end{equation}
\end{linenomath*}

Equation \eqref{eq:w_norm_1} combined with Equation \eqref{eq:w_0_1} implies that the vector \(\bmrm{w}\) must lie on a standard \(\paren{W - 1}\)-simplex, denoted as \(\bmrm{w} \in \Delta^{W - 1}\) hereafter, where
\begin{linenomath*}
\begin{equation}\label{eq:standard_simplex}
    \Delta ^{W - 1}=\left\{(t_{1},\dots ,t_{n})\in \mathbb {R} ^{W}~{\Bigg |}~\sum _{i=1}^{W}t_{i}=1{\text{ and }}t_{i}\geq 0{\text{ for }}i=1,\ldots ,W\right\}.
\end{equation}
\end{linenomath*}
Mathematically, this corresponds to the support of a Dirichlet distribution with order \(W\). Therefore, selecting the Dirichlet distribution as the prior for \(\bmrm{w}\) is a reasonable choice. Furthermore, in order to adopt a noninformative prior distribution that allows the data to determine the posterior distribution, we select a Dirichlet distribution with parameters \(\bmrm{1}_{W}\). This can be expressed as
\begin{linenomath*}
\begin{equation}
    \bmrm{w} \sim \operatorname{Dirichlet}\paren{\bmrm{1}_{W}}.
\end{equation}
\end{linenomath*}

Next, we need to determine the prior distribution of the UPs. The UP represents the probabilities of users associating themselves with different RPs when all the RPs are at full capacity. Similar to the constraint described in Equation \eqref{eq:w_norm_1}, the UP \(\bmrm{l}^{j}\) for each cluster \(j\) should lie within a standard \(\paren{N - 1}\)-simplex, as in Equation \eqref{eq:l_norm_1}
\begin{linenomath*}
\begin{equation}\label{eq:l_norm_1}
    \bmrm{l}^{j} \in \Delta^{N - 1}, j = 1, \ldots, W,
\end{equation}
\end{linenomath*}
where \(N\) denotes the total number of RPs being considered in the analysis.

On the other hand, the preference for each RP within each user cluster is assumed to be unknown \emph{a priori}, and a noninformative prior is desired for this distribution as well. Similar to the previous derivation, the Dirichlet distribution is chosen as the prior distribution for the UP, and it can be expressed as follows:
\begin{linenomath*}
\begin{equation}
    \bmrm{l}^{1}, \ldots, \bmrm{l}^{W} \sim \operatorname{Dirichlet}\paren{\bmrm{1}_{N}}.
\end{equation}
\end{linenomath*}
Note that the number of UPs should be equal to the number of clusters. 

For the sake of simplicity, the UPs \(\bmrm{l}^{1}, \ldots, \bmrm{l}^{W}\) can be stacked to form a UP matrix \(\bmrm{L}\), as shown in Equation \eqref{eq:stacked_L}
\begin{linenomath*}
\begin{equation}\label{eq:stacked_L}
    \bmrm{L} = \brackets{\bmrm{l}^{1}, \ldots, \bmrm{l}^{W}}^{\top} = \brackets{
    \begin{array}{cccc}
         \mathrm{l}^{1}_{1} & \mathrm{l}^{1}_{2} & \ldots & \mathrm{l}^{1}_{N} \\
         \mathrm{l}^{2}_{1} & \mathrm{l}^{2}_{2} & \ldots & \mathrm{l}^{2}_{N} \\
         \vdots & \vdots & \ddots & \vdots \\
         \mathrm{l}^{W}_{1} & \mathrm{l}^{W}_{2} & \ldots & \mathrm{l}^{W}_{N}
    \end{array}
    }
\end{equation}
\end{linenomath*}
The stacked UPs will also simplify the consequent computation, as described in the following paragraph.

Another important factor that affects the observed ULs associated with each RP is the explanatory variable, referred to as the RA of each RP. Denoted as \(\bmrm{u}\paren{k}\), the RA is a vector of length equal to the number of RPs, where each component represents the availability of the corresponding RP. To analyse the impact of \(\bmrm{u}\paren{k}\), we introduce the concept of a PSF for each UP-RA pair. The PSF denoted as \(\map{h}{\Delta^{N - 1} \times \left[0, +\infty\right)^{N}}{\Delta^{N - 1}}\), maps the RA and UP to a transformed PS. In this study, we adopt a simple yet reasonable choice for the PSF, which is given by:
\begin{linenomath*}
\begin{equation}\label{eq:PSF}
h\paren{\bmrm{l}^{j}, \bmrm{u}\paren{k}} = \frac{\bmrm{l}^{j} \odot \bmrm{u}\paren{k}}{\bmrm{l}^{j} \cdot \bmrm{u}\paren{k}},
\end{equation}
\end{linenomath*}
where \(\odot\) denotes element-wise multiplication between vectors. The chosen PSF captures the correlation between UP and RA, while ensuring that the PS vector is normalised and lies on a standard \(\paren{N - 1}\)-simplex.

To represent the impact of \(\bmrm{u}\paren{k}\) in a compact manner, we first construct a diagonal matrix whose elements are taken from \(\bmrm{u}\paren{k}\), as shown in Equation \eqref{eq:diagonal_u}
\begin{linenomath*}
\begin{equation}\label{eq:diagonal_u}
    \operatorname{diag}\paren{\bmrm{u}\paren{k}} = \brackets{
    \begin{array}{cccc}    
        \mathrm{u}_{1}\paren{k} & & & 0 \\
         & \mathrm{u}_{2}\paren{k} & & \\
         & & \ddots & \\
         0 & & & \mathrm{u}_{N}\paren{k}
    \end{array}
    },
\end{equation}
\end{linenomath*}
where \(\bmrm{u}\paren{k} = \brackets{\mathrm{u}_{1}\paren{k}, \mathrm{u}_{2}\paren{k}, \ldots, \mathrm{u}_{N}\paren{k}}^{\top}\). For simplicity, the time parameter \(k\) will be omitted from the components of \(\bmrm{u}\paren{k}\) in the following discussion.

To facilitate the computation of the numerator in Equation \eqref{eq:PSF}, we introduce an intermediate variable, \(\hat{\bmrm{A}}\paren{k + 1}\), as defined in Equation \eqref{eq:A^hat}:
\begin{linenomath*}
\begin{equation}\label{eq:A^hat}
    \hat{\bmrm{A}}\paren{k + 1} = \bmrm{L} \operatorname{diag}\paren{\bmrm{u}\paren{k}} = \brackets{
    \begin{array}{cccc}
         \mathrm{l}^{1}_{1} \mathrm{u}_{1} & \mathrm{l}^{1}_{2} \mathrm{u}_{2} & \ldots & \mathrm{l}^{1}_{N} \mathrm{u}_{N} \\
         \mathrm{l}^{2}_{1} \mathrm{u}_{1} & \mathrm{l}^{2}_{2} \mathrm{u}_{2} & \ldots & \mathrm{l}^{2}_{N} \mathrm{u}_{N} \\
         \vdots & \vdots & \ddots & \vdots \\
         \mathrm{l}^{W}_{1} \mathrm{u}_{1} & \mathrm{l}^{W}_{2} \mathrm{u}_{2} & \ldots & \mathrm{l}^{W}_{N} \mathrm{u}_{N}
    \end{array}
    },
\end{equation}
\end{linenomath*}
where the advancement of one time step signifies the temporal delay in the influence of \(\bmrm{u}\paren{k}\) upon \(\bmrm{x}\paren{k + 1}\).

To interpret the PS as the probability of each cluster associating with RPs, it is necessary to normalise \(\hat{\bmrm{A}}\paren{k + 1}\) so that it lies on a standard \(\paren{N - 1}\)-simplex. This can be achieved by computing the sum for each row in \(\hat{\bmrm{A}}\paren{k + 1}\), as illustrated in Equation \eqref{eq:A^hat_row_sum}:
\begin{linenomath*}
\begin{equation}\label{eq:A^hat_row_sum}
    \hat{\bmrm{A}}\paren{k + 1}\bmrm{1}_{N} = \brackets{
    \begin{array}{cccc}
         \mathrm{l}^{1}_{1} \mathrm{u}_{1} & \mathrm{l}^{1}_{2} \mathrm{u}_{2} & \ldots & \mathrm{l}^{1}_{N} \mathrm{u}_{N} \\
         \mathrm{l}^{2}_{1} \mathrm{u}_{1} & \mathrm{l}^{2}_{2} \mathrm{u}_{2} & \ldots & \mathrm{l}^{2}_{N} \mathrm{u}_{N} \\
         \vdots & \vdots & \ddots & \vdots \\
         \mathrm{l}^{W}_{1} \mathrm{u}_{1} & \mathrm{l}^{W}_{2} \mathrm{u}_{2} & \ldots & \mathrm{l}^{W}_{N} \mathrm{u}_{N}
    \end{array}
    } \brackets{
    \begin{array}{c}
         1 \\
         1 \\
         \vdots \\
         1
    \end{array}
    } = \brackets{
    \begin{array}{c}
         \mathrm{l}^{1}_{1} \mathrm{u}_{1} + \mathrm{l}^{1}_{2} \mathrm{u}_{2} + \ldots + \mathrm{l}^{1}_{N} \mathrm{u}_{N} \\
         \mathrm{l}^{2}_{1} \mathrm{u}_{1} + \mathrm{l}^{2}_{2} \mathrm{u}_{2} + \ldots + \mathrm{l}^{2}_{N} \mathrm{u}_{N} \\
         \vdots \\
         \mathrm{l}^{W}_{1} \mathrm{u}_{1} + \mathrm{l}^{W}_{2} \mathrm{u}_{2} + \ldots + \mathrm{l}^{W}_{N} \mathrm{u}_{N}
    \end{array}
    }.
\end{equation}
\end{linenomath*}
Next, we diagonalise the aforementioned row sum vector, as shown in Equation \eqref{eq:diagonal_row_sum}
\begin{linenomath*}
\begin{multline}\label{eq:diagonal_row_sum}
    \operatorname{diag}\paren{\hat{\bmrm{A}}\paren{k + 1}\bmrm{1}_{N}} = \\
    \brackets{
    \begin{array}{cccc}    
        \mathrm{l}^{1}_{1} \mathrm{u}_{1} + \mathrm{l}^{1}_{2} \mathrm{u}_{2} + \ldots + \mathrm{l}^{1}_{N} \mathrm{u}_{N} & & & 0 \\
         & \mathrm{l}^{2}_{1} \mathrm{u}_{1} + \mathrm{l}^{2}_{2} \mathrm{u}_{2} + \ldots + \mathrm{l}^{2}_{N} \mathrm{u}_{N} & & \\
         & & \ddots & \\
         0 & & & \mathrm{l}^{W}_{1} \mathrm{u}_{1} + \mathrm{l}^{W}_{2} \mathrm{u}_{2} + \ldots + \mathrm{l}^{W}_{N} \mathrm{u}_{N}
    \end{array}
    }.
\end{multline}
\end{linenomath*}
To serve as the normalisation term, the inverse matrix for the term in Equation \eqref{eq:diagonal_row_sum} should be calculated, as shown in Equation \eqref{eq:diagonal_row_sum_inv}
\begin{linenomath*}
\begin{multline}\label{eq:diagonal_row_sum_inv}
    \operatorname{diag}\paren{\hat{\bmrm{A}}\paren{k + 1}\bmrm{1}_{N}}^{-1} = \\
    \brackets{
    \begin{array}{cccc}    
        \frac{1}{\mathrm{l}^{1}_{1} \mathrm{u}_{1} + \mathrm{l}^{1}_{2} \mathrm{u}_{2} + \ldots + \mathrm{l}^{1}_{N} \mathrm{u}_{N}} & & & 0 \\
         & \frac{1}{\mathrm{l}^{2}_{1} \mathrm{u}_{1} + \mathrm{l}^{2}_{2} \mathrm{u}_{2} + \ldots + \mathrm{l}^{2}_{N} \mathrm{u}_{N}} & & \\
         & & \ddots & \\
         0 & & & \frac{1}{\mathrm{l}^{W}_{1} \mathrm{u}_{1} + \mathrm{l}^{W}_{2} \mathrm{u}_{2} + \ldots + \mathrm{l}^{W}_{N} \mathrm{u}_{N}}
    \end{array}
    }.
\end{multline}
\end{linenomath*}
Hence, we have obtained the normalisation matrix for \(\hat{\bmrm{A}}\paren{k + 1}\). Next, we should multiply the normalisation matrix to the left side of \(\hat{\bmrm{A}}\paren{k + 1}\), which gives rise to the PS matrix \(\bmrm{A}\paren{k + 1}\), as in Equation \eqref{eq:A^hat_normalized}
\begin{linenomath*}
\begin{multline}\label{eq:A^hat_normalized}
    \bmrm{A}\paren{k + 1} =  \operatorname{diag}\paren{\hat{\bmrm{A}}\paren{k + 1}\bmrm{1}_{N}}^{-1} \hat{\bmrm{A}}\paren{k + 1}  \\
    = \brackets{
    \begin{array}{cccc}
        \frac{\mathrm{l}^{1}_{1} \mathrm{u}_{1}}{\mathrm{l}^{1}_{1} \mathrm{u}_{1} + \mathrm{l}^{1}_{2} \mathrm{u}_{2} + \ldots + \mathrm{l}^{1}_{N} \mathrm{u}_{N}} & \frac{\mathrm{l}^{1}_{2} \mathrm{u}_{2}}{\mathrm{l}^{1}_{1} \mathrm{u}_{1} + \mathrm{l}^{1}_{2} \mathrm{u}_{2} + \ldots + \mathrm{l}^{1}_{N} \mathrm{u}_{N}} & \ldots & \frac{\mathrm{l}^{1}_{N} \mathrm{u}_{N}}{\mathrm{l}^{1}_{1} \mathrm{u}_{1} + \mathrm{l}^{1}_{2} \mathrm{u}_{2} + \ldots + \mathrm{l}^{1}_{N} \mathrm{u}_{N}} \\
        \frac{\mathrm{l}^{2}_{1} \mathrm{u}_{1}}{\mathrm{l}^{2}_{1} \mathrm{u}_{1} + \mathrm{l}^{2}_{2} \mathrm{u}_{2} + \ldots + \mathrm{l}^{2}_{N} \mathrm{u}_{N}} & \frac{\mathrm{l}^{2}_{2} \mathrm{u}_{2}}{\mathrm{l}^{2}_{1} \mathrm{u}_{1} + \mathrm{l}^{2}_{2} \mathrm{u}_{2} + \ldots + \mathrm{l}^{2}_{N} \mathrm{u}_{N}} & \ldots & \frac{\mathrm{l}^{2}_{N} \mathrm{u}_{N}}{\mathrm{l}^{2}_{1} \mathrm{u}_{1} + \mathrm{l}^{2}_{2} \mathrm{u}_{2} + \ldots + \mathrm{l}^{2}_{N} \mathrm{u}_{N}} \\
        \vdots & \vdots & \ddots & \vdots \\
        \frac{\mathrm{l}^{W}_{1} \mathrm{u}_{1}}{\mathrm{l}^{W}_{1} \mathrm{u}_{1} + \mathrm{l}^{W}_{2} \mathrm{u}_{2} + \ldots + \mathrm{l}^{W}_{N} \mathrm{u}_{N}} & \frac{\mathrm{l}^{W}_{2} \mathrm{u}_{2}}{\mathrm{l}^{W}_{1} \mathrm{u}_{1} + \mathrm{l}^{W}_{2} \mathrm{u}_{2} + \ldots + \mathrm{l}^{W}_{N} \mathrm{u}_{N}} & \ldots & \frac{\mathrm{l}^{W}_{N} \mathrm{u}_{N}}{\mathrm{l}^{W}_{1} \mathrm{u}_{1} + \mathrm{l}^{W}_{2} \mathrm{u}_{2} + \ldots + \mathrm{l}^{W}_{N} \mathrm{u}_{N}}
    \end{array}
    }.
\end{multline}
\end{linenomath*}

So far, we have obtained the PS for each cluster. However, to determine the observed ULs associated with each RP, we need to consider the weights of all the clusters. Mathematically, we calculate the cluster-weighted proportion vector \(\bmrm{p}\paren{k + 1}\) for the overall ULs associated with each RP. The vector \(\bmrm{p}\paren{k + 1}\) lies in the standard \(\paren{N - 1}\)-simplex and satisfies the constraint given by Equation \eqref{eq:p_simplex}:
\begin{linenomath*}
\begin{equation}\label{eq:p_simplex}
    \bmrm{p}\paren{k + 1} \in \Delta^{N - 1}.
\end{equation}
\end{linenomath*}
To determine \(\bmrm{p}\paren{k + 1}\), we should linearly combine the weight of each cluster, as shown in Equation \eqref{eq:p_combined}
\begin{linenomath*}
\begin{multline}\label{eq:p_combined}
    \bmrm{p}\paren{k + 1} = \paren{\bmrm{A}\paren{k + 1}}^{\top} \bmrm{w} \\
    = \brackets{
    \begin{array}{c}
        \frac{\mathrm{l}^{1}_{1} \mathrm{u}_{1} \mathrm{w}_{1}}{\mathrm{l}^{1}_{1} \mathrm{u}_{1} + \mathrm{l}^{1}_{2} \mathrm{u}_{2} + \ldots + \mathrm{l}^{1}_{N} \mathrm{u}_{N}} + \frac{\mathrm{l}^{2}_{1} \mathrm{u}_{1} \mathrm{w}_{2}}{\mathrm{l}^{2}_{1} \mathrm{u}_{1} + \mathrm{l}^{2}_{2} \mathrm{u}_{2} + \ldots + \mathrm{l}^{2}_{N} \mathrm{u}_{N}} + \ldots + \frac{\mathrm{l}^{\mathrm{w}}_{1} \mathrm{u}_{1} \mathrm{w}_{\mathrm{w}}}{\mathrm{l}^{\mathrm{w}}_{1} \mathrm{u}_{1} + \mathrm{l}^{\mathrm{w}}_{2} \mathrm{u}_{2} + \ldots + \mathrm{l}^{\mathrm{w}}_{N} \mathrm{u}_{N}} \\
        \frac{\mathrm{l}^{1}_{2} \mathrm{u}_{2} \mathrm{w}_{1}}{\mathrm{l}^{1}_{1} \mathrm{u}_{1} + \mathrm{l}^{1}_{2} \mathrm{u}_{2} + \ldots + \mathrm{l}^{1}_{N} \mathrm{u}_{N}} + \frac{\mathrm{l}^{2}_{2} \mathrm{u}_{2} \mathrm{w}_{2}}{\mathrm{l}^{2}_{1} \mathrm{u}_{1} + \mathrm{l}^{2}_{2} \mathrm{u}_{2} + \ldots + \mathrm{l}^{2}_{N} \mathrm{u}_{N}} + \ldots + \frac{\mathrm{l}^{\mathrm{w}}_{2} \mathrm{u}_{2} \mathrm{w}_{\mathrm{w}}}{\mathrm{l}^{\mathrm{w}}_{1} \mathrm{u}_{1} + \mathrm{l}^{\mathrm{w}}_{2} \mathrm{u}_{2} + \ldots + \mathrm{l}^{\mathrm{w}}_{N} \mathrm{u}_{N}} \\
        \vdots \\
        \frac{\mathrm{l}^{1}_{N} \mathrm{u}_{N} \mathrm{w}_{1}}{\mathrm{l}^{1}_{1} \mathrm{u}_{1} + \mathrm{l}^{1}_{2} \mathrm{u}_{2} + \ldots + \mathrm{l}^{1}_{N} \mathrm{u}_{N}} + \frac{\mathrm{l}^{2}_{N} \mathrm{u}_{N} \mathrm{w}_{2}}{\mathrm{l}^{2}_{1} \mathrm{u}_{1} + \mathrm{l}^{2}_{2} \mathrm{u}_{2} + \ldots + \mathrm{l}^{2}_{N} \mathrm{u}_{N}} + \ldots + \frac{\mathrm{l}^{\mathrm{w}}_{N} \mathrm{u}_{N} \mathrm{w}_{\mathrm{w}}}{\mathrm{l}^{\mathrm{w}}_{1} \mathrm{u}_{1} + \mathrm{l}^{\mathrm{w}}_{2} \mathrm{u}_{2} + \ldots + \mathrm{l}^{\mathrm{w}}_{N} \mathrm{u}_{N}}
    \end{array}
    }
\end{multline}
\end{linenomath*}
Thus, Equation \eqref{eq:p_closed} gives a compact form of \(\bmrm{p}\paren{k + 1}\),
\begin{linenomath*}
\begin{equation}\label{eq:p_closed}
\bmrm{p}\paren{k + 1} = \paren{\operatorname{diag}\paren{\bmrm{L} \operatorname{diag}\paren{\bmrm{u}\paren{k}} \bmrm{1}_{N}}^{-1} \bmrm{L} \operatorname{diag}\paren{\bmrm{u}\paren{k}}}^{\top} \bmrm{w}.
\end{equation}
\end{linenomath*}

It is assumed that the overall ULs at time \(k + 1\), denoted by \(M\) where \(M = \norm{\bmrm{x}\paren{k + 1}}_{1}\), can be inferred from the overall ULs at the preceding time step \(\norm{\bmrm{x}\paren{k}}_{1}\). Consequently, for the purpose of prediction, it becomes necessary only to delineate the proportion of ULs associated with each RP. The \(\bmrm{p}\paren{k + 1}\) furnishes insights into the mathematical expectation of the proportion of ULs that are associated with each RP within the existing RA. It is noteworthy, however, that the efficacy of the CLUSTER method is not solely determined by the expected value. It is also significantly influenced by fluctuations in the predicted proportion's variance. Such variability can introduce complexity into the model and affect the prediction's reliability. In order to accommodate this variance and to craft a more robust model, we utilise the Dirichlet distribution in conjunction with a concentration parameter, to represent the likelihood. This mathematical approach allows for greater flexibility and precision in modelling the underlying statistical properties of the ULs. Mathematically, the assumption can be expressed as
\begin{linenomath*}
\begin{equation}\label{eq:x_Dirichlet}
    \frac{\bmrm{x}\paren{k + 1}}{M} \sim \operatorname{Dirichlet}\paren{\mathrm{c}\bmrm{p}\paren{k + 1}},
\end{equation}
\end{linenomath*}
where \(\mathrm{c}\) serves as the concentration parameter and is also considered as a latent variable.

Thus, the complete structure of Na\"ive CLUSTER is given by Equation \eqref{eq:naive_CLUSTER}
\begin{linenomath*}
\begin{equation}\label{eq:naive_CLUSTER}
\begin{aligned}
    \frac{\bmrm{x}\paren{k + 1}}{M} &\sim \operatorname{Dirichlet}\paren{\mathrm{c}\bmrm{p}\paren{k + 1}}  \\
    \mathrm{c} &\sim \operatorname{Gamma}\paren{a, b} \\
    \bmrm{p}\paren{k + 1} &= \paren{\bmrm{A}\paren{k + 1}}^{\top} \bmrm{w}  \\
    \bmrm{A}\paren{k + 1} &= \operatorname{diag}\paren{\hat{\bmrm{A}}\paren{k + 1}\bmrm{1}_{N}}^{-1}\hat{\bmrm{A}}\paren{k + 1}  \\
    \hat{\bmrm{A}}\paren{k + 1} &= \bmrm{L}  \operatorname{diag}\paren{\bmrm{u}\paren{k}}  \\
    \bmrm{L} &= \brackets{\bmrm{l}^{1}, \ldots, \bmrm{l}^{W}}^{\top}  \\
    \bmrm{l}^{1}, \ldots, \bmrm{l}^{W} &\sim \operatorname{Dirichlet}\paren{\bmrm{1}_{N}}  \\
    \bmrm{w} &\sim \operatorname{Dirichlet}\paren{\bmrm{1}_{W}},
\end{aligned}
\end{equation}
\end{linenomath*}
where the parameters \(a\) and \(b\) are selected so that the prior distribution for \(\mathrm{c}\), \(\operatorname{Gamma}\paren{a, b}\), constitutes a weakly informative prior.

\newpage
\section{Integrating Dirichlet Process Mixture Model in Complete CLUSTER}

In this section, we delve into the structure of the Complete CLUSTER, which leverages the power of the Dirichlet process mixture model (DPMM). The integration of DPMM into our framework addresses a crucial challenge faced by Na\"ive CLUSTER: the manual presetting of the number of clusters, which inherently limits the flexibility of the model. By incorporating DPMM, we can automatically determine the appropriate number of clusters without the need for prior knowledge or manual intervention.

DPMM posits the existence of an infinite number of latent clusters, with each cluster assigned a weight that diminishes rapidly. However, from a practical standpoint, including an infinite number of components in the probabilistic model is infeasible due to computational constraints. In order to make CLUSTER computationally feasible, we employ the stick-breaking process and a truncation scheme to construct the components of the DPMM. This process requires the specification of a scaling parameter, denoted as \(\upalpha\), which determines the degree of discretisation of the Dirichlet process. A larger value of \(\upalpha\) indicates a smoother approximation.

To model the prior distribution of \(\upalpha\), we choose the Gamma distribution. The choice of the Gamma distribution is not unique, and alternative distributions with support on \(\left[0, +\infty\right)\) could also be utilised. The parameters of the Gamma distribution, \(c\) and \(d\), are chosen to make it a weakly informative prior, as described in Equation \eqref{eq:DPMM_alpha}:
\begin{linenomath*}
\begin{equation}\label{eq:DPMM_alpha}
\upalpha \sim \operatorname{Gamma}\paren{c, d}.
\end{equation}
\end{linenomath*}

It is important to note that the specific choice of the prior distribution parameters is a design decision and slight variations in these parameters are not expected to impact the inference and prediction results significantly. As the number of observations increases, the effect of the prior tends to diminish. Once the scaling parameter \(\upalpha\) is determined, the weights for each cluster are derived using the stick-breaking process, as illustrated in Equation \eqref{eq:stick-breaking}:
\begin{linenomath*}
\begin{equation}\label{eq:stick-breaking}
\begin{aligned}
    \upbeta_{1}, \ldots, \upbeta_{W} &\sim \operatorname{Beta}\paren{1, \upalpha}  \\
    \mathrm{w}_{j} &= \upbeta_{j} \prod_{k = j-1}^{j} \paren{1 - \upbeta_{k}}.
\end{aligned}
\end{equation}
\end{linenomath*}

Having obtained the weights for different clusters denoted as \(\bmrm{w}\), the remaining components of the Complete CLUSTER are identical to those of the Na\"ive CLUSTER, which are discussed in Section \ref{sec:naive_CLUSTER}. Therefore, we refrain from repeating the detailed explanation here. The full structure of Complete CLUSTER can then be formalised as in Equation \eqref{eq:full_CLUSTER}
\begin{linenomath*}
\begin{equation}\label{eq:full_CLUSTER}
\begin{aligned}
    \frac{\bmrm{x}\paren{k + 1}}{M} &\sim \operatorname{Dirichlet}\paren{\mathrm{c}\bmrm{p}\paren{k + 1}}  \\
    \mathrm{c} &\sim \operatorname{Gamma}\paren{a, b} \\
    \bmrm{p}\paren{k + 1} &= \paren{\bmrm{A}\paren{k + 1}}^{\top} \bmrm{w}  \\
    \bmrm{A}\paren{k + 1} &= \operatorname{diag}\paren{\hat{\bmrm{A}}\paren{k + 1}\bmrm{1}_{N}}^{-1}\hat{\bmrm{A}}\paren{k + 1}  \\
    \hat{\bmrm{A}}\paren{k + 1} &= \bmrm{L} \operatorname{diag}\paren{\bmrm{u}\paren{k}}  \\
    \bmrm{L} &= \brackets{\bmrm{l}^{1}, \ldots, \bmrm{l}^{W}}^{\top}  \\
    \bmrm{l}^{1}, \ldots, \bmrm{l}^{W} &\sim \operatorname{Dirichlet}\paren{\bmrm{1}_{N}}  \\
    \bmrm{w} &= \brackets{\mathrm{w}_{1}, \ldots, \mathrm{w}_{W}}^{\top}  \\
    \mathrm{w}_{j} &= \upbeta_{j} \prod_{k = j-1}^{j} \paren{1 - \upbeta_{k}}  \\
    \upbeta_{1}, \ldots, \upbeta_{W} &\sim \operatorname{Beta}\paren{1, \upalpha}  \\
    \upalpha &\sim \operatorname{Gamma}\paren{c, d}.
\end{aligned}
\end{equation}
\end{linenomath*}

\newpage
\section{Modelling Deterministic Selection in User Preferences}

The introduction of UPs in CLUSTER is a direct consequence of acknowledging the inherent stochastic nature of ULs in association with various RPs. It has been observed that the selection made by individual users often encapsulates a probabilistic pattern, reflecting the variability and uncertainty embedded within their preferences. However, in some application scenarios, users may display deterministic patterns in their RP selection process. This deterministic behaviour is characterised by the selection of a specific RP until its availability falls below a predetermined threshold, upon which the stochastic nature of the decision-making process resurfaces. In this section, we endeavour to demonstrate that our CLUSTER framework can effectively adapt to and capture such deterministic settings, delineating underlying patterns of UPs and their association with RPs.

Consider a scenario in which each user has a distinct order of preferences for RPs. To encapsulate this order, we employ a vector notation defined in Equation \eqref{eq:preference_vector_order}:
\begin{linenomath*}
\begin{equation}\label{eq:preference_vector_order}
    \bmrm{o}^{j} = \brackets{\mathrm{o}^{j}_{1}, \mathrm{o}^{j}_{2}, \ldots, \mathrm{o}^{j}_{N}}^{\top},
\end{equation}
\end{linenomath*}
where \(\bmrm{o}^{j}\) denotes the UP order vector for the \(j\)-th user. Each element of the vector corresponds to a unique RP ID number, and the sequence of elements reflects the user's UP order, with the first element being the most preferred RP. Crucially, \(\bmrm{o}^{j}\) signifies a permutation of the entire set of possible RP IDs. Thus, every possible RP ordering is valid, and \(\bmrm{o}^{j}\) is a member of the set of all permutations of \(\until{N}\).

Next, we define \(\mathrm{l}^{j}_{i}\) as the \(i\)-th element of the UP vector for the \(j\)-th user. The \(\bmrm{o}^{j}\)-indexed vector, \(\bmrm{c}^{j} = \brackets{\mathrm{l}^{j}_{\mathrm{o}^{j}_{1}}, \ldots, \mathrm{l}^{j}_{\mathrm{o}^{j}_{N}}}^{\top}\), represents the ordered UP vector for the \(j\)-th user. To incorporate the above-mentioned application scenario, we set \(\bmrm{r}^{j} \in \left[1, +\infty\right)^{N-1}\), and the \(i\)-th component of \(\bmrm{r}^{j}\) is defined as per Equation \eqref{eq:large_k}:
\begin{linenomath*}
\begin{equation}\label{eq:large_k}
    \mathrm{r}_{i}^{j} = \frac{\mathrm{c}^{j}_{i}}{\mathrm{c}^{j}_{i + 1}}, i = 1, \ldots, N - 1.
\end{equation}
\end{linenomath*}
The ratio \(\mathrm{r}_{i}^{j}\) depicts the relative UP of the user towards the preceding RP, vis-à-vis the next proximate alternative in the ordered UP vector. A large \(\mathrm{r}_{i}^{j}\) value can be interpreted as an almost deterministic choice. 

Hence, the deterministic characteristics of the ULs can be appropriately captured by judicious selection of the \(\bmrm{r}^{j}\) vector, thereby seamlessly integrating the deterministic setting into CLUSTER. In practical application, the vector \(\bmrm{c}^{j}\) can be inferred employing MCMC sampling techniques.

\newpage
\section{Navigating High-Dimensional Data with t-SNE in User Preference Visualisation}

In this section, we elucidate our approach for visualising the UPs. One of the challenges their high dimensionality. Considering that the number of RPs, denoted as \(N\), is typically a large number, direct visualisation becomes untenable. Hence, it is essential to deploy a technique capable of mapping high-dimensional data into a lower-dimensional space. One such potent technique is t-distributed stochastic neighbour embedding (t-SNE), a machine learning algorithm explicitly designed for the visualisation and dimensionality reduction of high-dimensional data. t-SNE excels at transforming these data into two or three dimensions, making it particularly suitable for graphical representation. The primary objective of t-SNE is to derive a faithful representation of high-dimensional points in a lower-dimensional space, typically within the confines of a 2D plane or 3D space. Consequently, for this study, we adopt the t-SNE technique to elucidate the patterns within the UPs effectively.

Our chief aim is to scrutinise the UPs produced by tests that use a progressively decreasing number of preset clusters. These tests make use of the simpler version of our model, known as the Na\"ive CLUSTER. To get a full understanding of the hidden variables, we must bring together the results from various sets of samples, all produced by the MCMC samplers. However, if we tried to map all of these vectors, we would quickly find ourselves swamped with data. The sheer number of UPs would make the graph incredibly dense and therefore harder to understand. To maintain a clear visualisation while ensuring that we include a comprehensive set of MCMC samples, we opt to select a large but limited number of UPs from a subset of the MCMC samples. To keep our visuals consistent across different experiments, we make sure to select an equal number of vectors, no matter the number of preset clusters in the experiment. This consistent approach allows us to better compare results from different conditions, making our findings more reliable.

Our analysis involves mapping the \(N\)-dimensional UPs to both two-dimensional planes and three-dimensional spaces, to gain a more comprehensive understanding of the underlying structures. In the case of two dimensions, the visualisation process seeks to underscore the existence of distinct clusters. To accomplish this, we adopt a colour-coding strategy, wherein points belonging to disparate clusters are represented with different colours. Notably, our aim is not only to indicate the presence of clear-cut clusters but also to represent the degree of uncertainty regarding the points that lie in the transitional regions between clusters. In the context of our visualisations, these are points that may not unequivocally belong to one cluster or another. We use gradient colours to achieve this, which allows us to illustrate the nuanced uncertainties in these transitional regions. To attain this more sophisticated visual representation, we implement a weighted colour determination process based on each point's distance from the centre of each cluster. This process gives more importance to a cluster as the point gets closer to its centre, reflecting its stronger association. Consequently, points located in between different clusters will exhibit a mix of colours that corresponds to their relative distances from the centres of the surrounding clusters. This approach allows us to generate insightful visualisations that portray the intricacies of the UPs in a clear and intuitive manner.

\newpage
\section*{Supplementary Figures}

\begin{figure*}[t]  

\centering   \includegraphics[width=0.925\linewidth]{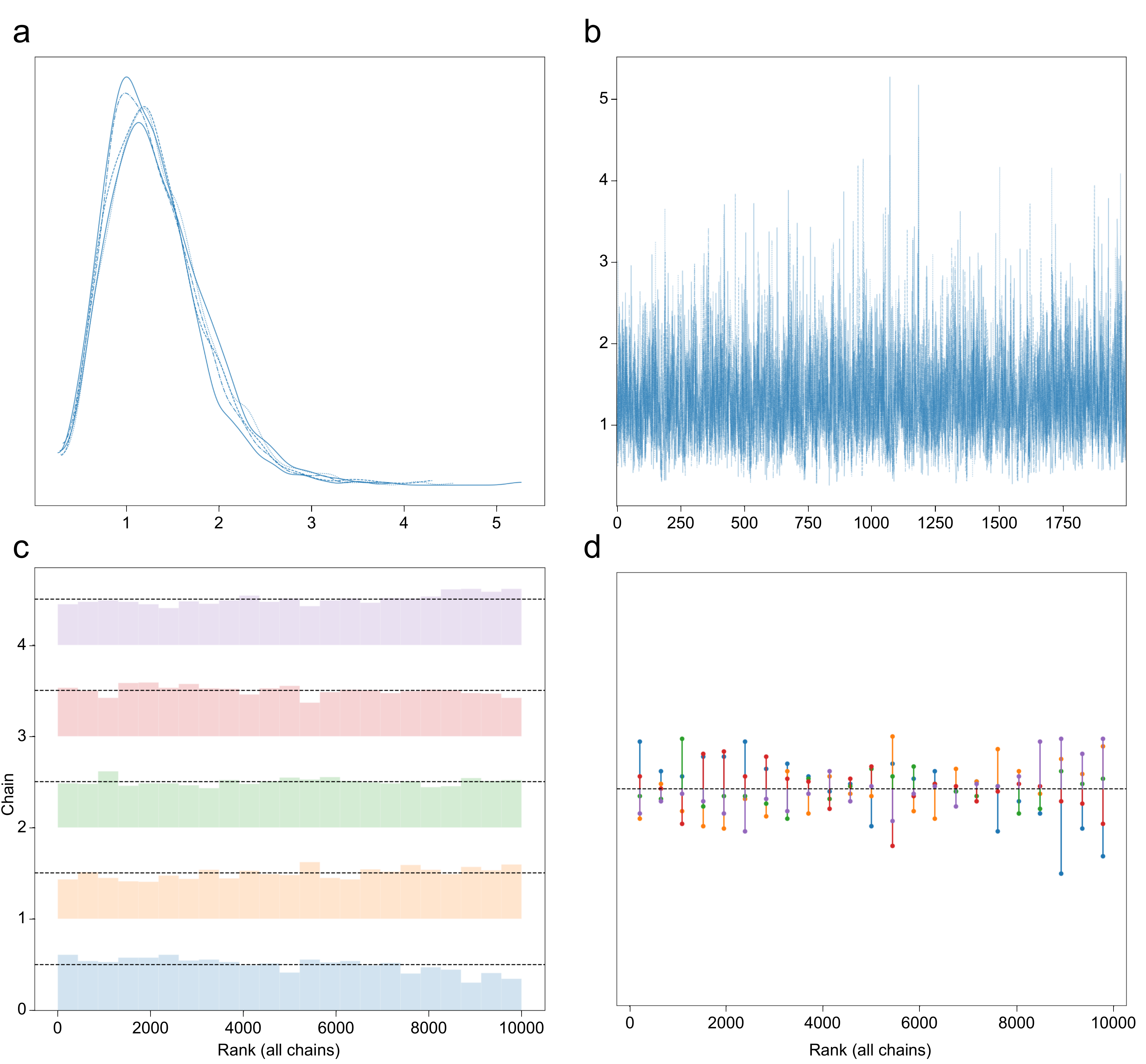}  \vspace{0.5cm} \caption{\linespread{1}\selectfont{}\textbf{Posterior plots of the scaling parameter \(\upalpha\) in Complete CLUSTER for simulated data analysis.} Panel \textbf{a} delineates the posterior distribution of \(\upalpha\), ascertained through the individual utilisation of Markov chains employing MCMC samplers. Within these distributions, depicted via kernel density estimation (KDE), there is a predominant concentration within the interval \(\brackets{0.5, 2}\) across all respective chains, thereby indicating satisfactory convergence. In panel \textbf{b}, trace plots for each chain are displayed. These plots, which are consistent with the distribution patterns seen in panel \textbf{a}, primarily cluster around values ranging from 0.5 to 2, thereby affirming the chains' convergence. Panels \textbf{c} and \textbf{d} expound the rank order statistics of the chains. Under ideal conditions, the ranks within each chain would adhere to a uniform distribution. The bar plot exhibited in panel \textbf{c} manifests an alignment closely paralleling a theoretical uniform distribution, thereby signifying the convergence of all chains to an identical posterior distribution. Panel \textbf{d} furnishes an in-depth analysis of the discrepancies from the uniform distribution observed in panel \textbf{c}, thereby contributing further understanding into potential irregularities within the chains.
}
\label{fig:Synthetic_DPMM_alpha_trace}
\end{figure*}

\begin{figure*}[t]  

\centering   \includegraphics[width=0.925\linewidth]{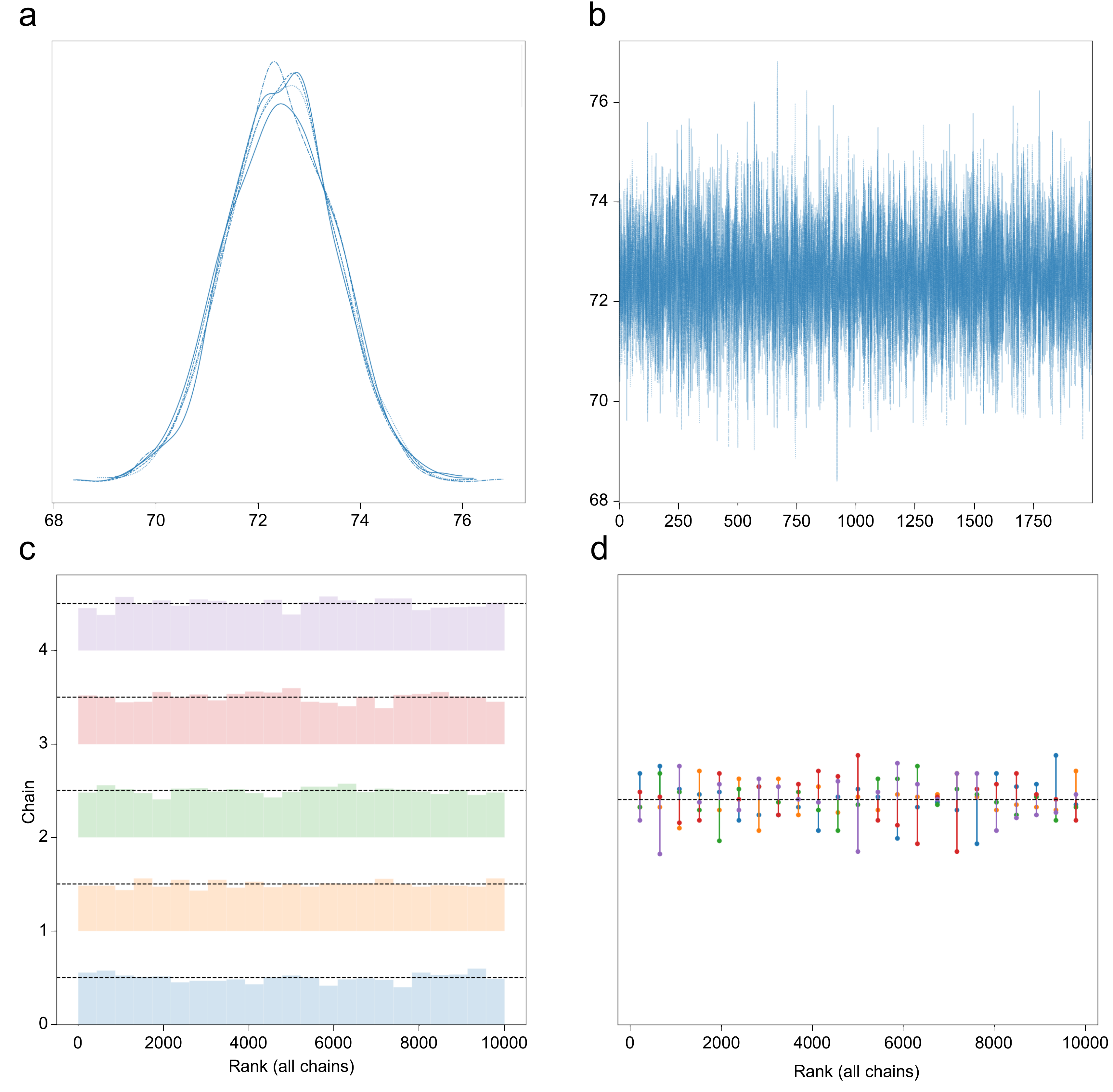}  \vspace{0.5cm} \caption{\linespread{1}\selectfont{}\textbf{Posterior plots of the concentration parameter \(\mathrm{c}\) in Complete CLUSTER for simulated data analysis.} Panel \textbf{a} delineates the posterior distribution of \(\mathrm{c}\), ascertained through the individual utilisation of Markov chains employing MCMC samplers. Within these distributions, depicted via KDE, there is a predominant concentration within the interval \(\brackets{70, 74}\) across all respective chains, thereby indicating satisfactory convergence. In panel \textbf{b}, trace plots for each chain are displayed. These plots, which are consistent with the distribution patterns seen in panel \textbf{a}, primarily cluster around values ranging from 70 to 74, thereby affirming the chains' convergence. Panels \textbf{c} and \textbf{d} expound the rank order statistics of the chains. The bar plot exhibited in panel \textbf{c} manifests an alignment closely paralleling a theoretical uniform distribution, thereby signifying the convergence of all chains to an identical posterior distribution. Panel \textbf{d} furnishes an in-depth analysis of the discrepancies from the uniform distribution observed in panel \textbf{c}, thereby contributing further understanding into potential irregularities within the chains.
}
\label{fig:Synthetic_DPMM_c_trace}
\end{figure*}

\begin{figure*}[t]  

\centering   \includegraphics[width=0.925\linewidth]{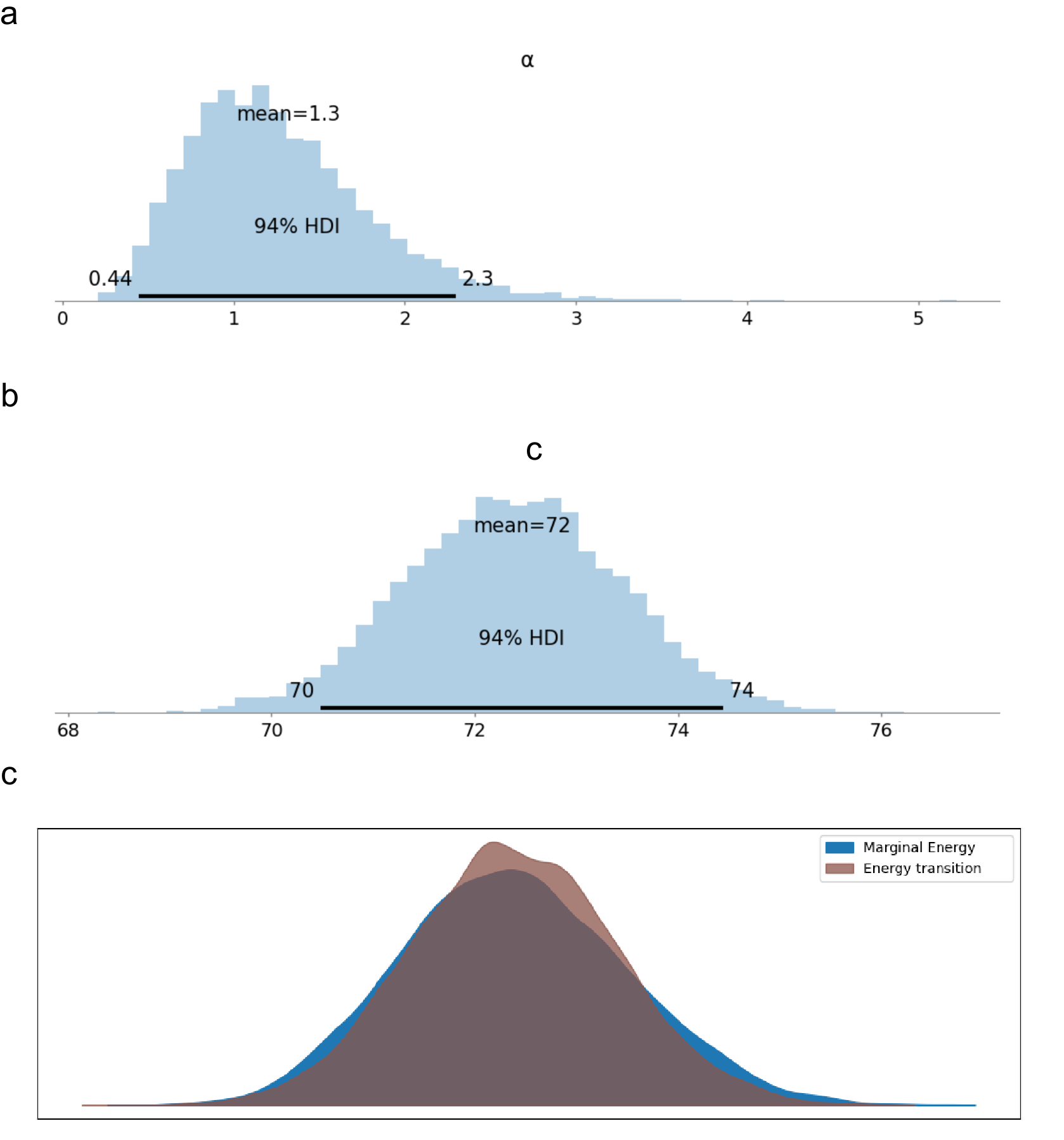}  \vspace{0.5cm} \caption{\linespread{1}\selectfont{}\textbf{Histogram for the posterior distribution of the scaling parameter \(\upalpha\) and concentration parameter \(\mathrm{c}\) in Complete CLUSTER for simulated data analysis and the energy plot.} In panel \textbf{a}, we visualise the posterior distribution of the scaling parameter \(\upalpha\) by combining all MCMC samples from five separate Markov chains. This histogram reveals a mean \(\upalpha\) value of approximately 1.3, encapsulated within a \(94\%\) Highest Density Interval (HDI) ranging from 0.44 to 2.3. In panel \textbf{b}, we visualise the posterior distribution of the concentration parameter \(\mathrm{c}\). This histogram reveals a mean \(\mathrm{c}\) value of approximately 72, encapsulated within a \(94\%\) HDI ranging from 70 to 74. Panel \textbf{c} presents the energy plot, a crucial tool in evaluating the performance of the No-U-Turn Sampler (NUTS) algorithm employed in our analyses. An energy plot ensures adequate exploration of the posterior distribution; insufficient exploration risks biased estimates due to infrequent visits to certain sections of the posterior. The large overlapping area observed in this panel suggests satisfactory exploration, enhancing the reliability of our posterior inference.
}
\label{fig:Synthetic_DPMM_alpha_posterior_energy}
\end{figure*}

\begin{figure*}[t]  

\centering   \includegraphics[width=0.925\linewidth]{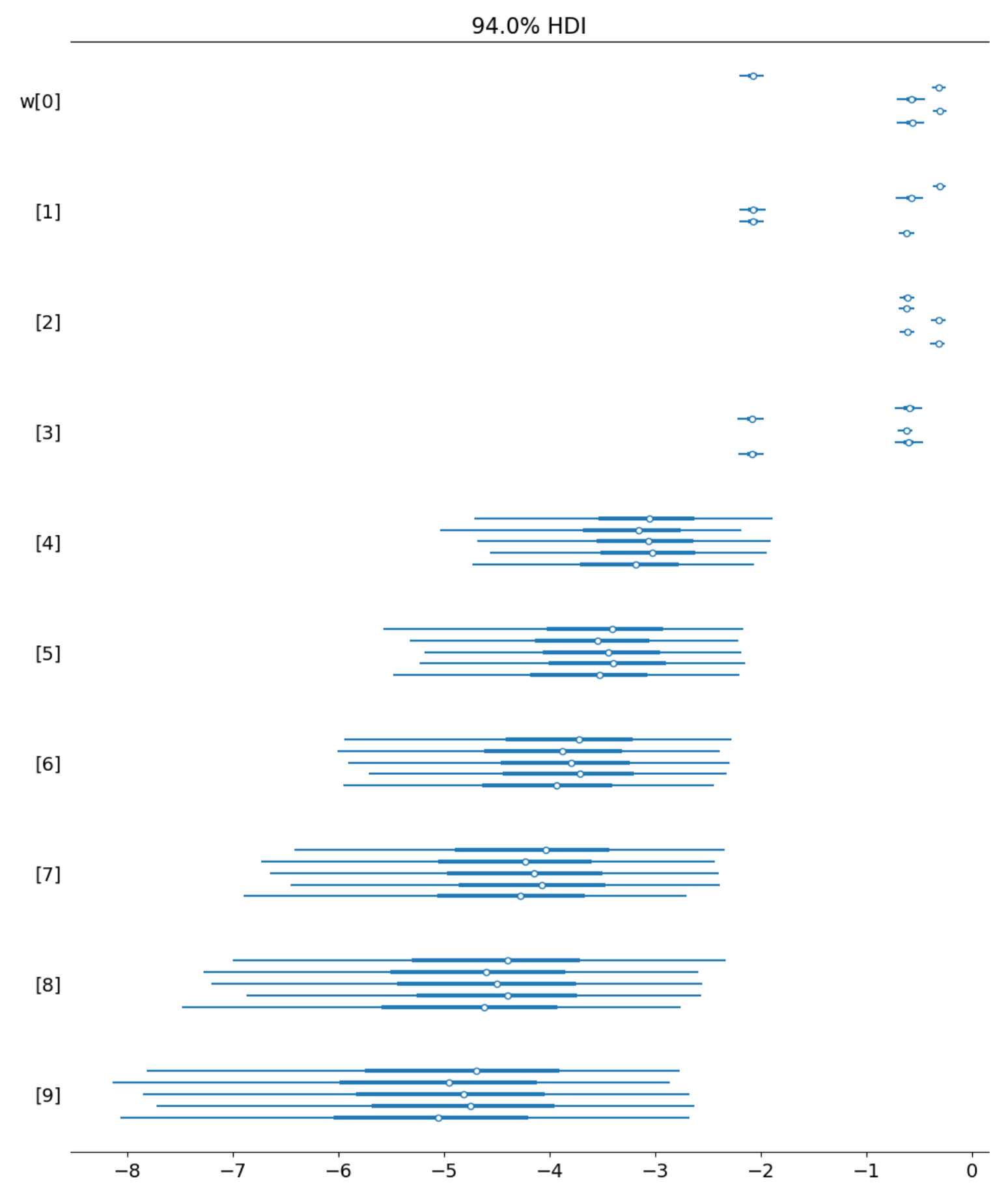}  \vspace{0.5cm} \caption{\linespread{1}\selectfont{}\textbf{Forest plot for the posterior distribution of the weight vector \(\bmrm{w}\) in Complete CLUSTER for simulated data analysis.} This figure presents a forest plot of the posterior distribution for the weight vector \(\bmrm{w}\) within the Complete CLUSTER, which incorporates samples from all five Markov chains. Each horizontal line represents samples from an individual chain. For visual clarity, we display only ten components of the weight vector in the logarithmic scale (\(\operatorname{log}_{10}\)). Interestingly, the plot reveals that as the number of clusters increases, the weight corresponding to the newly introduced cluster decreases rapidly. This observation lends support to our truncation strategy.
}
\label{fig:Synthetic_DPMM_w_forest}
\end{figure*}

\begin{figure*}[t]  

\centering   \includegraphics[width=0.925\linewidth]{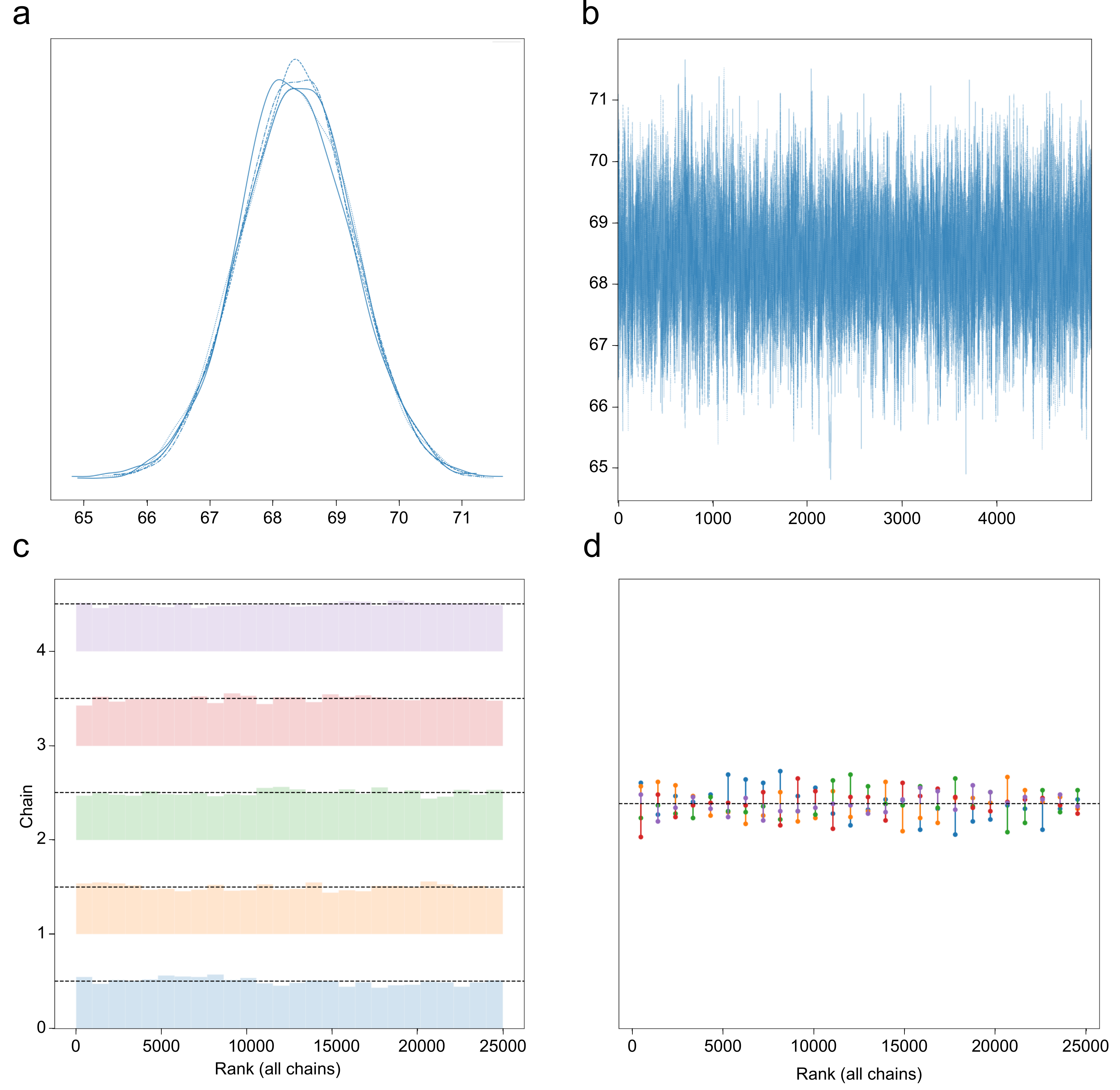}  \vspace{0.5cm} \caption{\linespread{1}\selectfont{}\textbf{Posterior plots of the concentration parameter \(\mathrm{c}\) in Na\"ive CLUSTER for simulated data analysis.} We manually set the number of clusters to 10 in the following experiments to evaluate the performance of Na\"ive CLUSTER. Panel \textbf{a} delineates the posterior distribution of \(\mathrm{c}\), ascertained through the individual utilisation of Markov chains employing MCMC samplers. Within these distributions, depicted via KDE, there is a predominant concentration within the interval \(\brackets{67, 70}\) across all respective chains, thereby indicating satisfactory convergence. In panel \textbf{b}, trace plots for each chain are displayed. These plots, which are consistent with the distribution patterns seen in Panel \textbf{a}, primarily cluster around values ranging from 67 to 70, thereby affirming the chains' convergence. Panels \textbf{c} and \textbf{d} expound the rank order statistics of the chains. The bar plot exhibited in panel \textbf{c} manifests an alignment closely paralleling a theoretical uniform distribution, thereby signifying the convergence of all chains to an identical posterior distribution. Panel \textbf{d} furnishes an in-depth analysis of the discrepancies from the uniform distribution observed in panel \textbf{c}, thereby contributing further understanding into potential irregularities within the chains.
}
\label{fig:Synthetic_naive_c_trace}
\end{figure*}

\begin{figure*}[t]  

\centering   \includegraphics[width=0.925\linewidth]{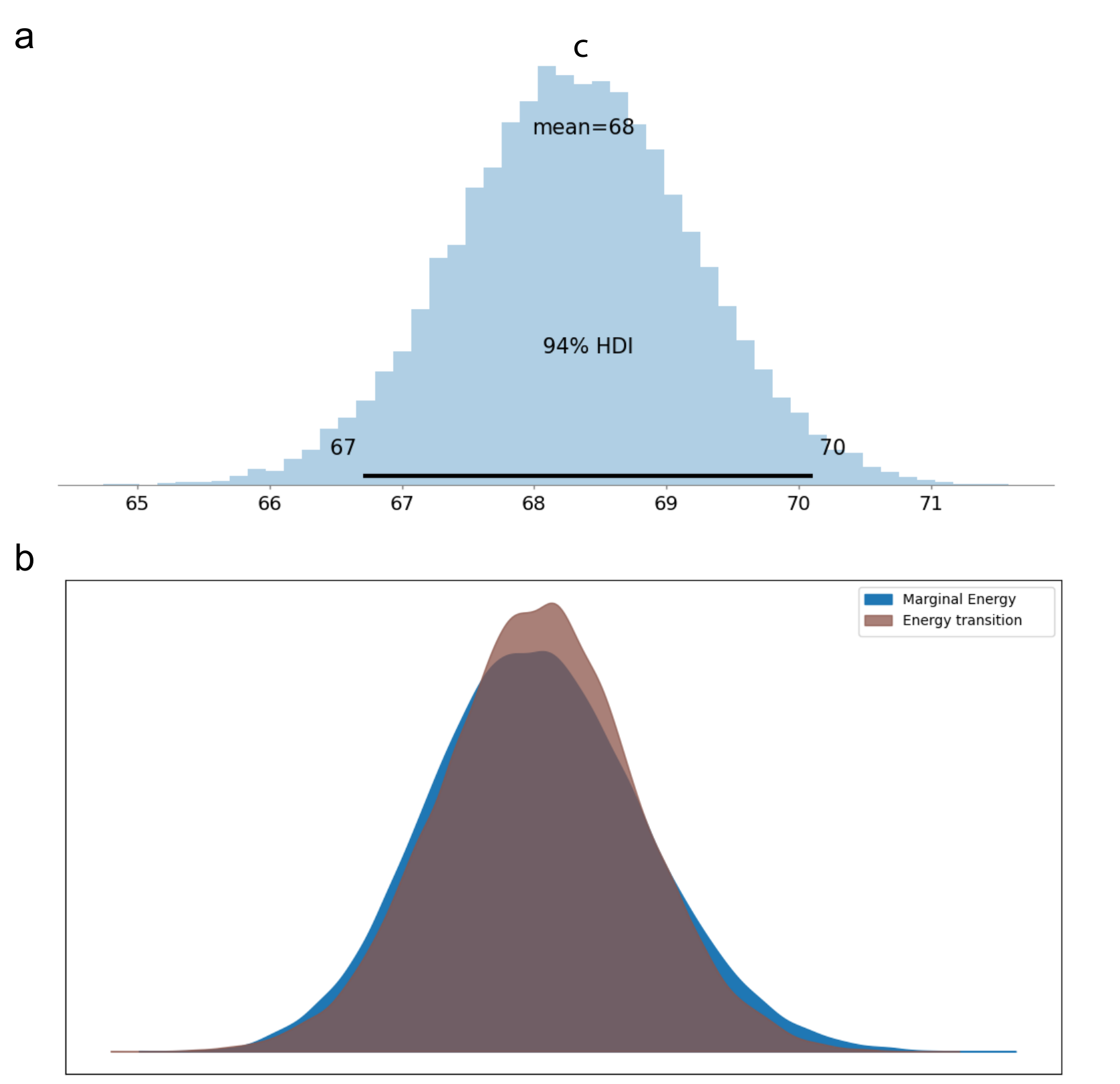}  \vspace{0.5cm} \caption{\linespread{1}\selectfont{}\textbf{Histogram for the posterior distribution of the concentration parameter \(\mathrm{c}\) in Na\"ive CLUSTER for simulated data analysis and the energy plot.} In panel \textbf{a}, we visualise the posterior distribution of the concentration parameter \(\mathrm{c}\) by combining all MCMC samples from five separate Markov chains. This histogram reveals a mean \(\mathrm{c}\) value of approximately 68, encapsulated within a \(94\%\) HDI ranging from 67 to 70. Panel \textbf{b} presents the energy plot. The large overlapping area observed in this panel suggests satisfactory exploration, enhancing the reliability of our posterior inference.
}
\label{fig:Synthetic_naive_c_posterior_energy}
\end{figure*}

\begin{figure*}[t]  

\centering   \includegraphics[width=0.925\linewidth]{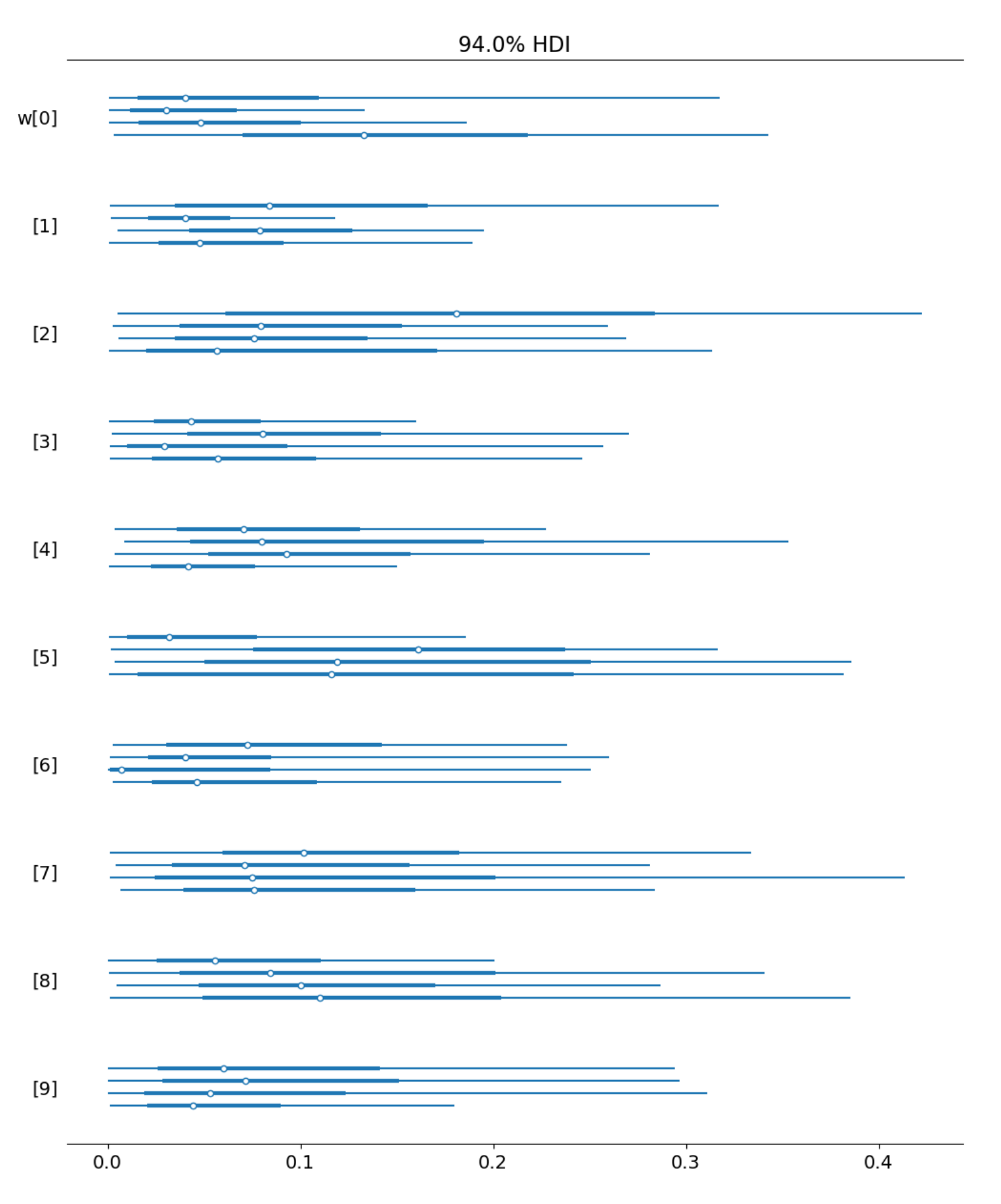}  \vspace{0.5cm} \caption{\linespread{1}\selectfont{}\textbf{Forest plot for the posterior distribution of the weight vector \(\bmrm{w}\) in Na\"ive CLUSTER for simulated data analysis.} This figure presents a forest plot of the posterior distribution for the weight vector \(\bmrm{w}\) within the Na\"ive CLUSTER, which incorporates samples from all four Markov chains. Each horizontal line represents samples from an individual chain. In this experiment, we intentionally fix the number of clusters to 10 as a design choice. Given the inherent exchangeability of the components within the latent weight vector, the posterior distribution for each component demonstrates a minimal level of deviation from one another. Predominantly, most of the components converge approximately around the value of 0.1. 
}
\label{fig:Synthetic_Naive_10_w_forest}
\end{figure*}

\begin{figure*}[t]  

\centering   \includegraphics[width=0.925\linewidth]{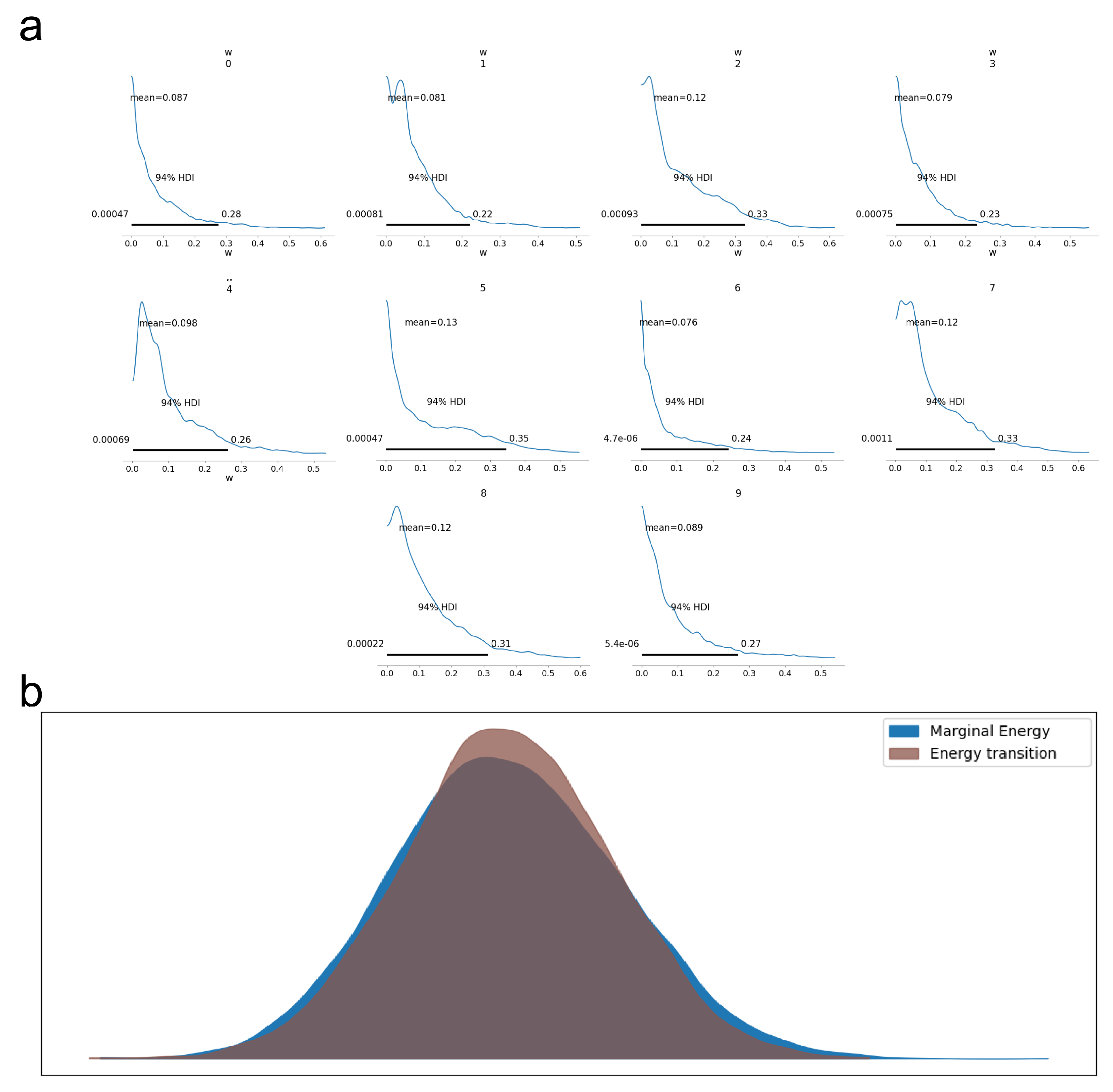}  \vspace{0.5cm} \caption{\linespread{1}\selectfont{}\textbf{Posterior distribution of the weight vector \(\bmrm{w}\) and the energy plot in Na\"ive CLUSTER for simulated data analysis.} The panel \textbf{a} in the figure illustrates the posterior distribution for each component of the weight vector. Here, KDE is used for visualisation, with samples drawn from different Markov chains combined for this purpose. Notably, the mean values of the weights strongly suggest the exchangeability characteristic intrinsic to these components, as most of them cluster closely around the value of 0.1. In panel \textbf{b}, the energy plot is depicted. It showcases two bell-shaped curves, with their intersection illustrating a large overlapping area, thereby affirming that our sampler has conducted a comprehensive exploration of the posterior distribution.
}
\label{fig:Synthetic_Naive_10_w_combine}
\end{figure*}

\begin{figure*}[t]  

\centering   \includegraphics[width=1\linewidth]{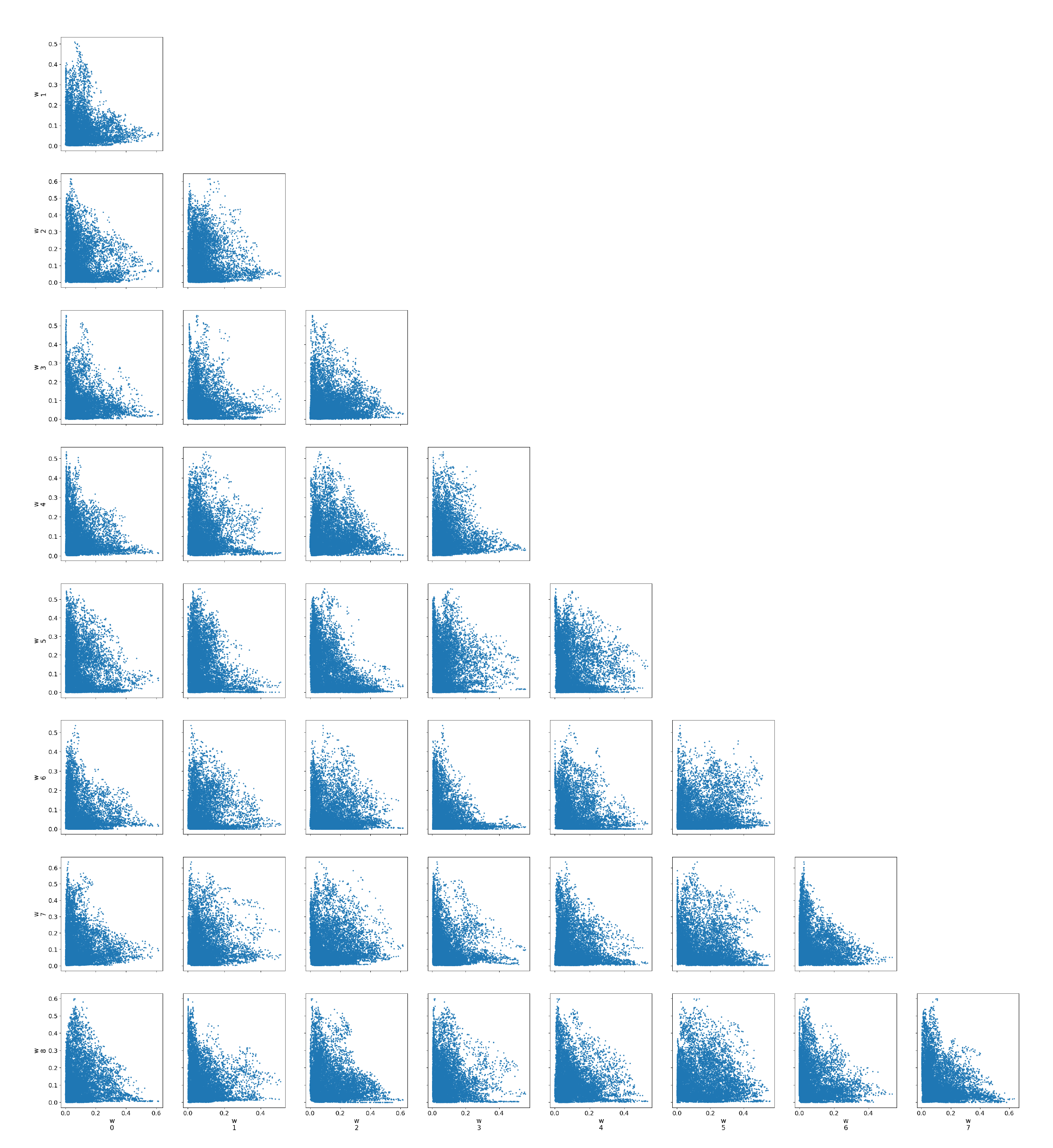}  \vspace{0.5cm} \caption{\linespread{1}\selectfont{}\textbf{Pair plot of the weight vector \(\bmrm{w}\) in Na\"ive CLUSTER for simulated data analysis.} This figure presents the scatter plots for the simulated samples from the MCMC sampler of the weight vector, \(\bmrm{w}\). Here, each point corresponds to a sample, with the x and y coordinates representing distinct components within the vector. On a broad scale, it is observed that most of the samples across all components tend to cluster around the value of 0.1. More interestingly, an evident correlation can be seen between these components. This correlation is a direct consequence of the inherent constraint imposed on the components of \(\bmrm{w}\), which is the condition that their summation must equal 1.  
}
\label{fig:Synthetic_Naive_10_w_pair}
\end{figure*}

\begin{figure*}[t]  

\centering   \includegraphics[width=0.925\linewidth]{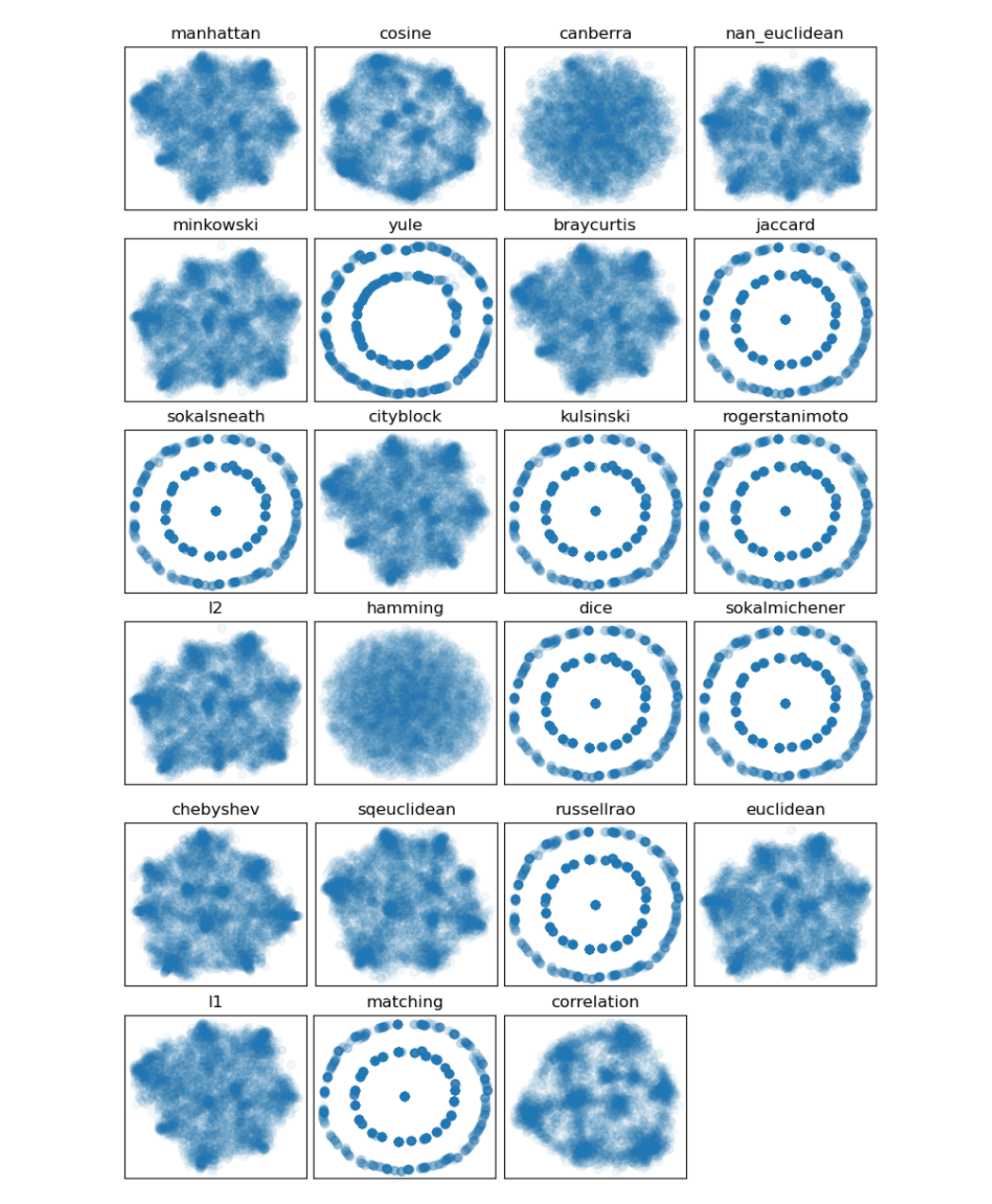}  \vspace{0.5cm} \caption{\linespread{1}\selectfont{}\textbf{Different distance metrics for t-SNE visualisation.} This figure illustrates the outcome of applying t-SNE to map high-dimensional UPs to a 2-dimensional plane. A set of 10,000 UP samples is employed for this visualisation, and they are evaluated using an assortment of 23 distinct metrics. Our analysis shows that the metric that computes the correlation between UPs in the high-dimensional space gives the most insightful visualisation.
}
\label{fig:Synthetic_Naive_tsne_metric}
\end{figure*}

\begin{figure*}[t]  

\centering   \includegraphics[width=0.925\linewidth]{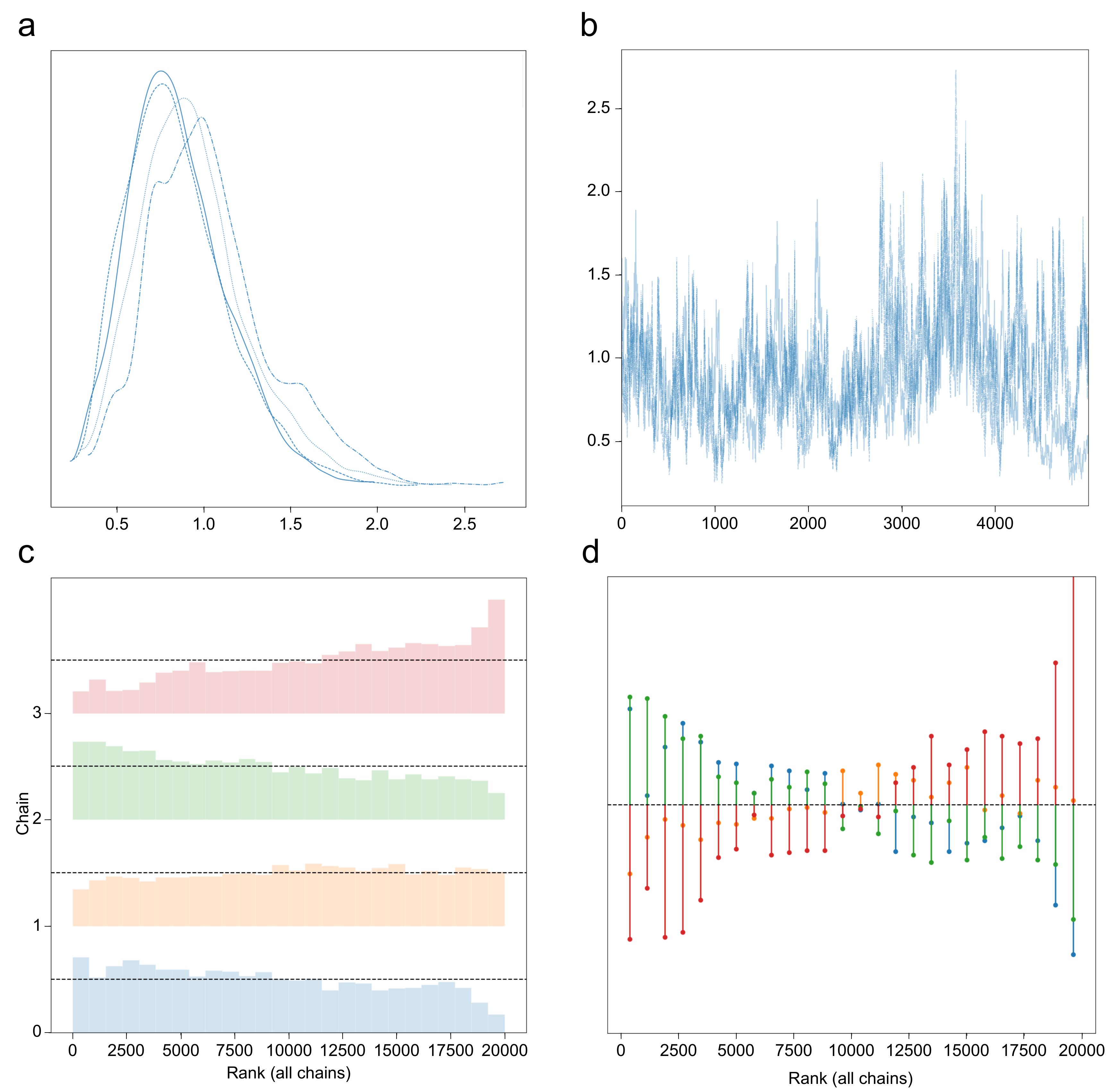}  \vspace{0.5cm} \caption{\linespread{1}\selectfont{}\textbf{Posterior plots of the scaling parameter \(\upalpha\) in Complete CLUSTER for real-world base station (BS) data analysis.} Panel \textbf{a} showcases the distribution of \(\upalpha\), as determined through four separate Markov chains using MCMC samplers. The distributions largely concentrate within the interval \(\brackets{0.35, 1.5}\) across all chains, suggesting good convergence. Panel \textbf{b} displays trace plots for each chain. These plots, mirroring the distribution findings in panel \textbf{a}, mostly centre around values between 0.35 and 1.5, affirming the convergence of the chains. Panel \textbf{c} and panel \textbf{d} illustrate the rank order statistics of chains. The bar plot in panel \textbf{c} reveals a close alignment with a theoretical uniform distribution, indicating the convergence of all chains to the same posterior distribution. Panel \textbf{d} provides a detailed analysis of the deviations from the uniform distribution seen in panel \textbf{c}, offering further insights into potential anomalies within the chains.
}
\label{fig:Realistic_DPMM_alpha_trace}
\end{figure*}

\begin{figure*}[t]  

\centering   \includegraphics[width=0.925\linewidth]{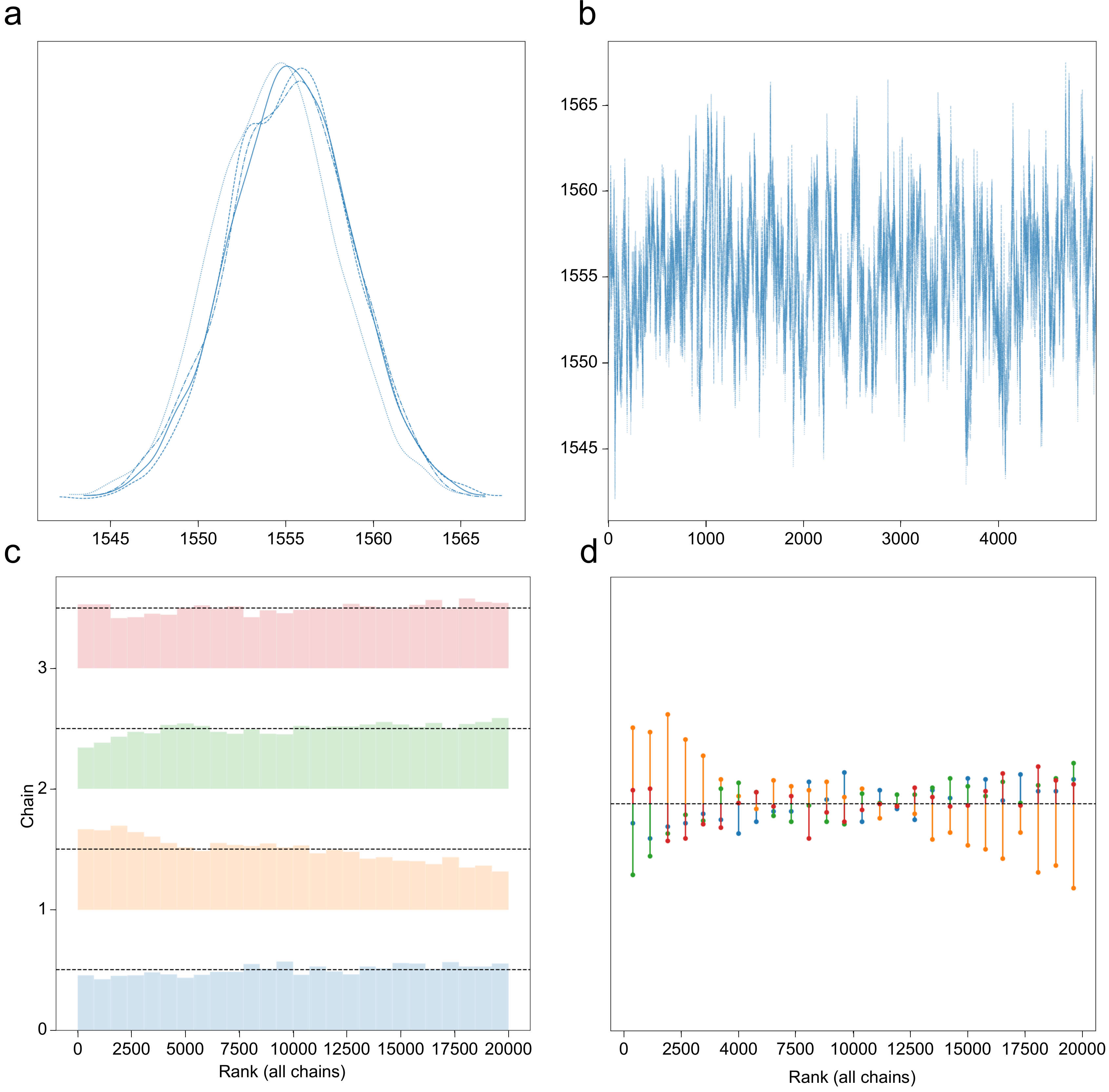}  \vspace{0.5cm} \caption{\linespread{1}\selectfont{}\textbf{Posterior plots of the concentration parameter \(\mathrm{c}\) in Complete CLUSTER for real-world BS data analysis.} Panel \textbf{a} showcases the distribution of \(\mathrm{c}\), as determined through four separate Markov chains using MCMC samplers. The distributions largely concentrate within the interval \(\brackets{1548, 1562}\) across all chains, suggesting good convergence. Panel \textbf{b} displays trace plots for each chain. These plots, mirroring the distribution findings in panel \textbf{a}, mostly centre around values between 1548 and 1562, affirming the convergence of the chains. Panel \textbf{c} and panel \textbf{d} illustrate the rank order statistics of chains. The bar plot in panel \textbf{c} reveals a close alignment with a theoretical uniform distribution, indicating the convergence of all chains to the same posterior distribution. Panel \textbf{d} provides a detailed analysis of the deviations from the uniform distribution seen in panel \textbf{c}, offering further insights into potential anomalies within the chains.
}
\label{fig:Realistic_DPMM_c_trace}
\end{figure*}

\begin{figure*}[t]  

\centering   \includegraphics[width=0.925\linewidth]{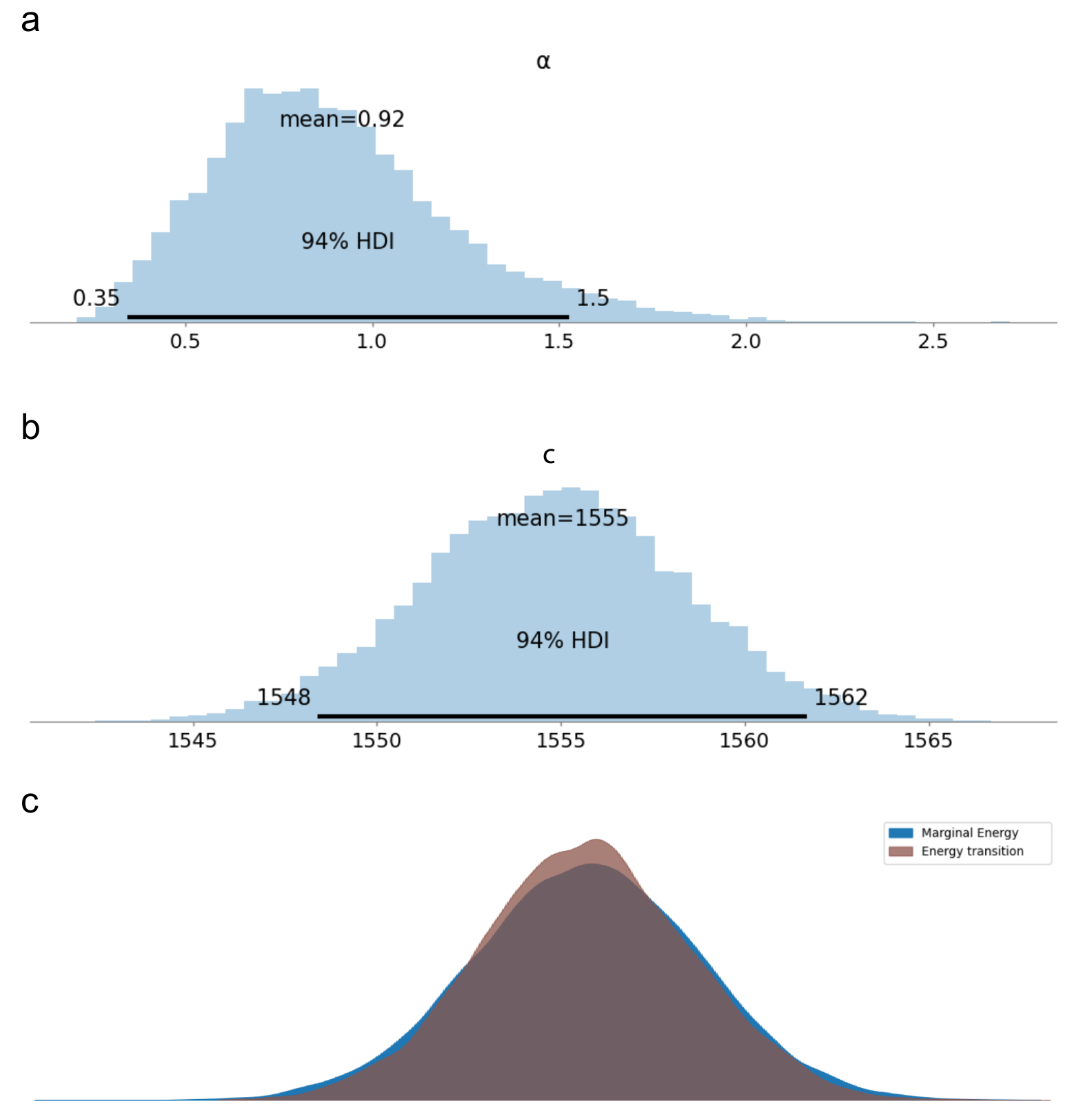}  \vspace{0.5cm} \caption{\linespread{1}\selectfont{}\textbf{Histogram for the posterior distribution of the scaling parameter \(\upalpha\) and concentration parameter \(\mathrm{c}\) in Complete CLUSTER for real-world BS data analysis and the energy plot.} In Panel \textbf{a}, we visualise the posterior distribution of the scaling parameter \(\upalpha\) by combining all MCMC samples from five separate Markov chains. This histogram reveals a mean \(\upalpha\) value of approximately 0.92, encapsulated within a \(94\%\) HDI ranging from 0.35 to 1.5. In Panel \textbf{b}, we visualise the posterior distribution of the concentration parameter \(\mathrm{c}\). This histogram reveals a mean \(\mathrm{c}\) value of approximately 1555, encapsulated within a \(94\%\) HDI ranging from 1548 to 1562. Panel \textbf{c} presents the energy plot. The large overlapping area observed in this panel suggests satisfactory exploration, enhancing the reliability of our posterior inference.
}
\label{fig:Realistic_DPMM_alpha_posterior_energy}
\end{figure*}

\begin{figure*}[t]  

\centering   \includegraphics[width=0.8\linewidth]{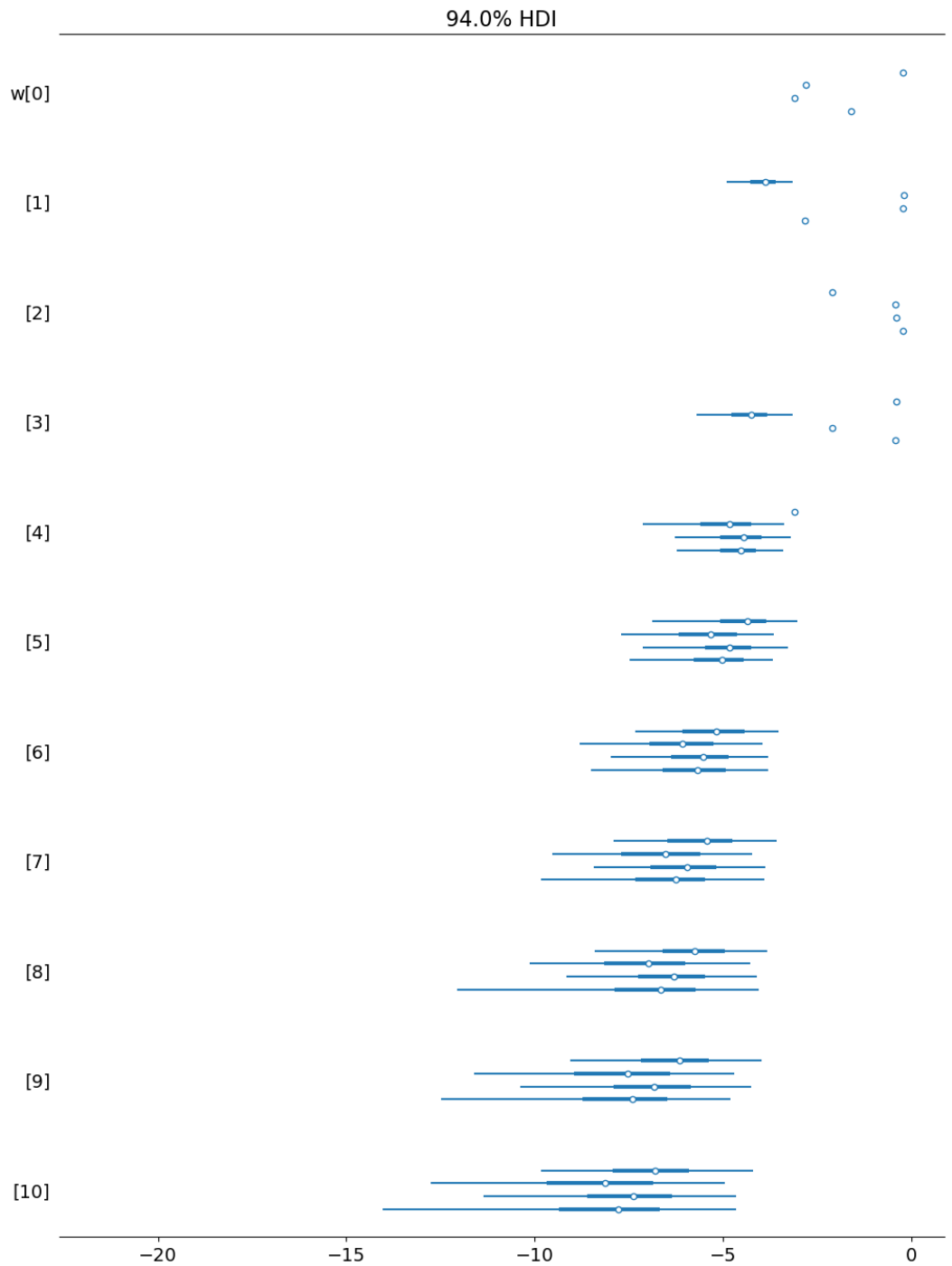}  \vspace{0.5cm} \caption{\linespread{1}\selectfont{}\textbf{Forest plot for the posterior distribution of the weight vector \(\bmrm{w}\) in Complete CLUSTER for real-world BS data analysis.} This figure presents a forest plot of the posterior distribution for the weight vector \(\bmrm{w}\) within the Complete CLUSTER, which incorporates samples from all four Markov chains. Each horizontal line represents samples from an individual chain. For visual clarity, we display only ten components of the weight vector in the logarithmic scale (\(\operatorname{log}_{10}\)).
}
\label{fig:Realistic_DPMM_10_w_forest}
\end{figure*}

\begin{figure*}[t]  

\centering   \includegraphics[width=0.925\linewidth]{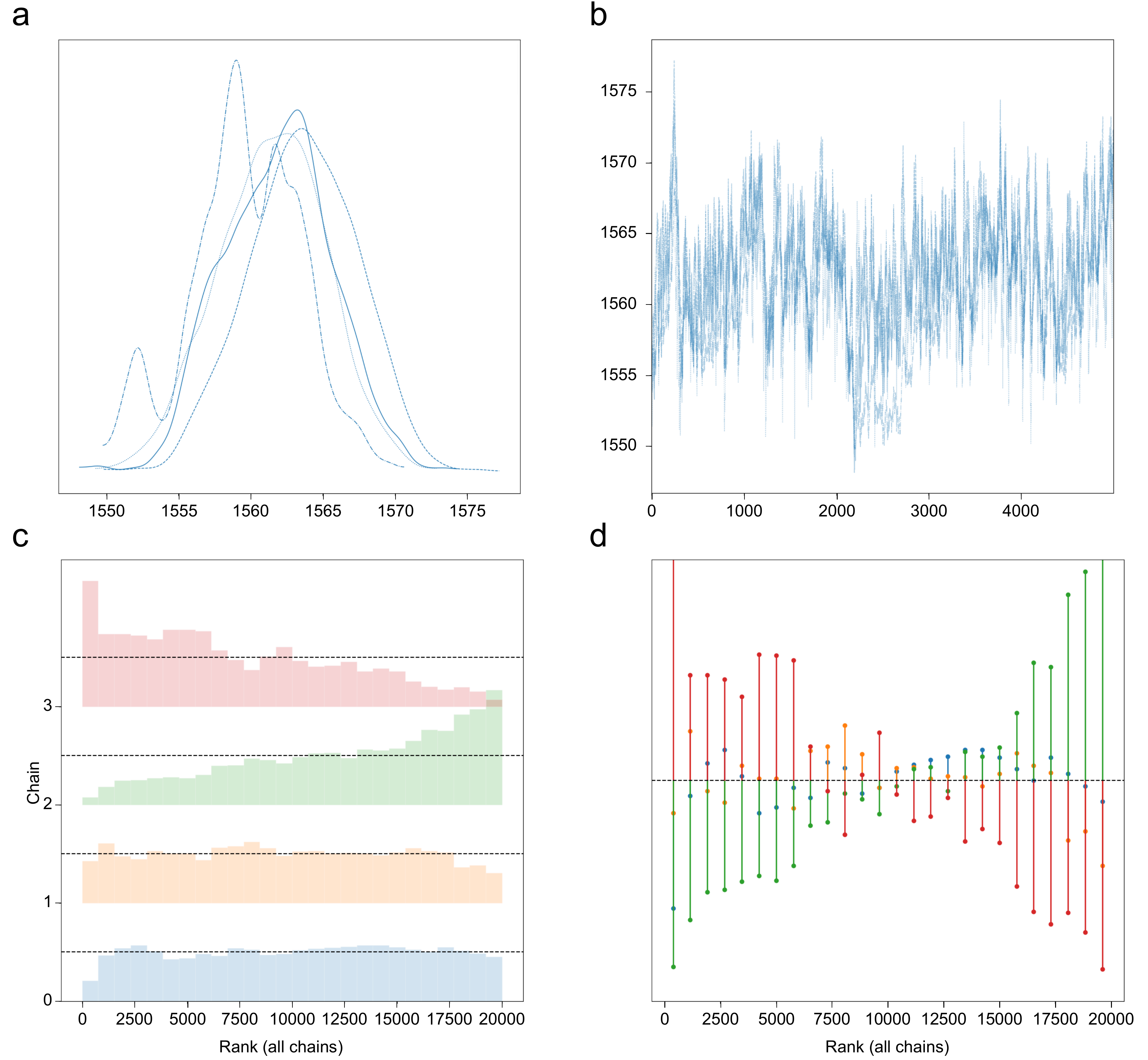}  \vspace{0.5cm} \caption{\linespread{1}\selectfont{}\textbf{Posterior plots of the concentration parameter \(\mathrm{c}\) in Complete CLUSTER for real-world BS data analysis.} Panel \textbf{a} showcases the distribution of \(\mathrm{c}\), as determined through four separate Markov chains using MCMC samplers. The distributions largely concentrate within the interval \(\brackets{1548, 1562}\) across all chains, suggesting good convergence. Panel \textbf{b} displays trace plots for each chain. These plots, mirroring the distribution findings in panel \textbf{a}, mostly centre around values between 1548 and 1562, affirming the convergence of the chains. Panel \textbf{c} and panel \textbf{d} illustrate the rank order statistics of chains. The bar plot in panel \textbf{c} reveals a close alignment with a theoretical uniform distribution, indicating the convergence of all chains to the same posterior distribution. Panel \textbf{d} provides a detailed analysis of the deviations from the uniform distribution seen in panel \textbf{c}, offering further insights into potential anomalies within the chains.
}
\label{fig:Realistic_Naive_c_trace}
\end{figure*}

\newpage
\bibliography{alias,FP,Main,New}